\documentclass{article}

\usepackage[table, dvipsnames]{xcolor}
\usepackage[preprint]{corl_2026} 

\usepackage{fontawesome5}

\usepackage{amsmath}
\usepackage{amsfonts}
\usepackage{mathtools}     
\usepackage[makeroom]{cancel}

\usepackage{enumitem}
\usepackage{graphicx}
\usepackage{wrapfig}
\usepackage{subfig}
\usepackage{mwe}
\usepackage{microtype}
\usepackage{booktabs}
\usepackage{multirow}
\usepackage{makecell}

\usepackage{algorithm}
\usepackage{algpseudocode}

\usepackage{multicol}

\usepackage{svg}

\usepackage{siunitx}
\usepackage{placeins}

\usepackage{pifont}
\newcommand{\cmark}{\textcolor{green!60!black}{\checkmark}} 
\newcommand{\xmark}{\textcolor{red}{\ding{55}}}

\definecolor{gainblue}{RGB}{26,134,163}
\definecolor{gainred}{RGB}{215,95,76}

\newcommand{\gainup}[1]{\textcolor{gainblue}{#1\textsubscript{$\uparrow$}}}
\newcommand{\gaindown}[1]{\textcolor{gainred}{#1\textsubscript{$\downarrow$}}}

\usepackage{capt-of} 

\usepackage{ifthen}
\newboolean{reftomain}
\setboolean{reftomain}{true}

\begin{document}

\title{Attention from Action, for Action: \\Emergent Visual Bottlenecks for Policy Learning}

\author{
Zheyu Zhuang$^{1}$
\href{mailto:roboticist.zheyu@gmail.com}
{\textcolor{black}{\raisebox{0.1ex}{\scriptsize\faEnvelope[regular]}}}
\hspace{0.5pt},
Ruiyu Wang$^{1}$,
Nick Heppert$^{2, 4}$,
Johannes Fabian Hahn$^{3}$,\\
\textbf{
Abhinav Valada$^{2}$,
Florian T. Pokorny$^{1}$,
Danica Kragic$^{1}$}\\[0.35em]
\small $^{1}$Department of Robotics, Perception and Learning, KTH Royal Institute of Technology, Sweden\\
\small $^{2}$Department of Computer Science, University of Freiburg, Germany\\
\small $^{3}$Department of Informatics, Universität Hamburg, Germany\\
\small $^{4}$ZUSE School ELIZA
}

\hypersetup{
  pdftitle={Attention from Action, for Action: Emergent Visual Bottlenecks for Policy Learning},
  pdfauthor={Zheyu Zhuang, Ruiyu Wang, Nick Heppert, Johannes Fabian Hahn, Abhinav Valada, Florian T. Pokorny, Danica Kragic}
}
\maketitle

\begin{abstract}
Visual bottlenecks that focus policy inputs on regions of interest (ROIs) can improve the data efficiency of visuomotor learning by separating \emph{where to look} from \emph{how to act}.
However, where to look changes as a task progresses, particularly in long-horizon manipulation, where the evidence needed for control shifts among objects, contact regions, and target locations.
Existing ROI interfaces often rely on external spatial cues, while action-derived alternatives center crops on the projected tool center point at keyframes identified by gripper or motion transitions.
These predefined temporal and spatial heuristics can miss visual evidence needed for the next action.
To address these limitations, we introduce Seeker, which learns where to look through action prediction, without spatial annotations or prescribed keyframes.
A task- and state-conditioned query iteratively gathers evidence from frozen DINOv3 features, producing adaptive ROIs for RGB cropping, mask-guided background augmentation, and point-cloud filtering.
Across MimicGen tasks with $100$ demonstrations per task, Seeker improves average success from $42.6\%$ for the strongest input-level baseline to $62.6\%$.
On real robots, average success rises from $48.3\%$ to $76.7\%$ in-domain and from $20.0\%$ to $60.0\%$ under challenging lighting and background shifts.
\href{https://github.com/zheyu-zhuang/seeker}{%
\raisebox{-0.15ex}{\faGithub}\,\textsc{Code}}
\end{abstract}

\keywords{Imitation Learning, Data Efficiency, Region of Interest (ROI)}

\section{Introduction}
Imitation learning is applied to long-horizon manipulation in visually complex environments, where decision-critical cues are sparse and localized~\cite{wang2023mimicplay}.
A way to improve data efficiency is to separate \emph{where to look} from \emph{how to act}: estimate a region of interest (ROI) and train the policy on the resulting focused observation~\cite{takizawa2025enhancing,james2022coarse_to_fine,wang2026palm}.
The open challenge is that the ROI is a control-dependent visual bottleneck rather than a fixed semantic category.
It can shift with robot state, task phase, and camera view, and may correspond to an object, an affordance region, a tool--object contact, a goal region, or a spatial relation between objects.
Existing ROI interfaces often rely on spatial cues or semantic priors, including predicted human gaze~\cite{takizawa2025enhancing}, end-effector-centered crops~\cite{wang2026palm}, and vision--language model grounding~\cite{liu2024grounding,moo2023openworld,kwon2024language}.
These cues may not specify the evidence needed for the next control decision; grounding long-horizon tasks can require stage-aware prompts or subtask decomposition~\cite{hu2026clap,myers2024policy}.

The readily available action stream, however, provides a control-aligned learning signal for where to look. Action-grounded heuristics exploit this connection by deriving spatial targets from demonstration trajectories, for example by projecting the end-effector at keyframes identified from motion stops or gripper-state changes~\cite{james2022coarse_to_fine,shridhar2023perceiver,goyal2024rvt2}. While avoiding spatial annotations, these methods depend on hand-designed rules for when and where to attend: discrete motion events may be absent or task-irrelevant, and fixed-size crops around the end-effector may exclude important contextual cues.

We propose Seeker to learn task- and state-conditioned spatial selection through action prediction, without spatial annotations or predefined trajectory events and proxy points.
Seeker initializes a query from the task and robot state, then iteratively updates it using visual evidence retrieved from frozen DINOv3~\cite{dinov3_2025} patch features.
The resulting attention defines an adaptive ROI whose location and extent depend on the current observation.
Once trained, Seeker is frozen and provides ROIs for downstream use.
Simulation and real-world evaluations show:
\begin{enumerate}[label=\textbf{(\arabic*)},leftmargin=*,itemsep=0pt,topsep=1pt,parsep=0pt,partopsep=0pt]
    \item \textbf{Action supervision yields reusable, policy-useful ROIs:}
    Seeker approaches the privileged Oracle ROI reference in RGB policy learning. The same frozen module also supports mask-guided background augmentation and simulated point-cloud filtering.
    \item \textbf{The learned bottleneck improves data efficiency:}
    with $100$ demonstrations per task and a shared downstream RGB stack, Seeker raises average simulation success from $42.6\%$ to $62.6\%$. It also outperforms a strong full-image policy baseline at $1{,}000$ demonstrations per task.
    \item \textbf{Learned ROIs improve real-world robustness:}
    Seeker leads the baselines across three hardware tasks in-domain and under lighting and background changes, where its average success is three times that of the strongest baseline ($60.0\%$ vs.~$20.0\%$).
\end{enumerate}

\section{Related Work}

\paragraph{Spatial Guidance and Active Vision.}
Given fixed camera views, policies have selected task-relevant regions using predicted human gaze~\cite{takizawa2025enhancing}, VLM grounding~\cite{liu2024grounding,moo2023openworld}, or simply the projected end-effector location, as in PALM~\cite{wang2026palm}. PALM centers its crops on the end-effector, assuming that nearby image regions contain the cues needed for interaction. When the camera is controllable, spatial selection extends to choosing the view to acquire.
AV-ALOHA~\cite{chuang2025avaloha} and Vision in Action~\cite{xiong2025visioninaction} learn camera movement from demonstrations, while Look, Focus, Act~\cite{chuang2026lookfocusact} uses human gaze to guide foveated processing and EyeRobot~\cite{kerr2025eyerobot} learns camera gaze from an action-prediction reward.

\paragraph{Action-Derived Spatial Priors.}
Manipulation trajectories offer a different source of localization targets.
A common strategy identifies keyframes from gripper-state changes or low end-effector velocity and uses the next gripper pose as a spatial target.
Q-attention~\cite{james2022q_attention} projects this pose into the image to guide learning of a crop, while Coarse-to-Fine Q-attention~\cite{james2022coarse_to_fine} refines a predicted translation through successive voxel grids.
PerAct~\cite{shridhar2023perceiver} predicts keyframe poses from voxelized scenes; RVT~\cite{goyal2023rvt} and RVT-2~\cite{goyal2024rvt2} use multi-view representations, with RVT-2 rendering closer views around a predicted region of interest.
For these methods, the location is predicted at inference, but the chosen event and gripper-pose proxy determine its training target.
CLAP~\cite{hu2026clap} builds on RVT-2's coarse-to-fine design and instead guides fine-grained prediction with subtask language and VLM-predicted 3D keypoints.

\paragraph{Spatial Structure without ROI Selection.}
Geometric structure can also guide visuomotor learning without defining a localization target.
MirrorDuo~\cite{mirrorduo2025} jointly reflects images, robot states, and actions so that their geometric correspondence is preserved.
Equivariant Diffusion Policy~\cite{wang2024equivariant} builds rotational equivariance into action generation.
RAVEN~\cite{klee2026raven} represents RGB observations as 3D rays to construct an SE(3)-equivariant policy.
These methods place geometric structure in the training data or policy, leaving the selection of task-dependent visual evidence implicit.

\section{Methodology}
\label{sec:method}
\begin{figure}[t]
\centering
\includegraphics[width=0.88\linewidth]{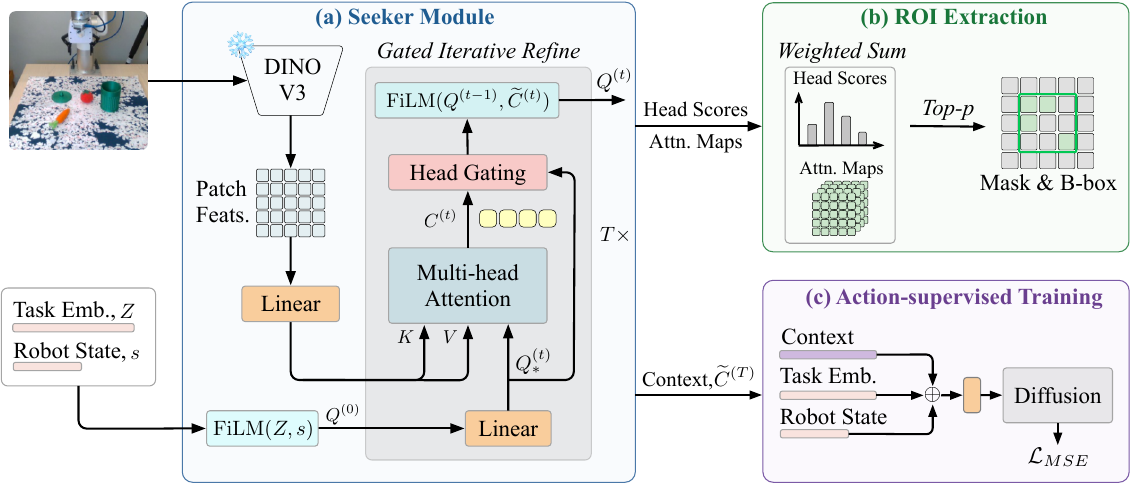}
\vspace{-1mm}
\caption{
\textbf{Seeker Overview.}
(a) A state-conditioned query is iteratively refined over frozen DINOv3 patch tokens, producing a compact visual context together with attention maps and head scores.
(b) Head-score-weighted attention is converted into a mask and bounding box using a \texttt{top\_p} mass criterion.
(c) The visual context is trained through a diffusion action-prediction loss, allowing the ROI to emerge without spatial supervision.
}
\label{fig:seeker_overview}
\vspace{-3mm}
\end{figure}

Fig.~\ref{fig:seeker_overview}(a) illustrates Seeker, a lightweight
readout over frozen DINOv3~\cite{dinov3_2025} patch tokens.
Seeker adapts attention-probing readouts~\cite{bardes2024cv_attentive_probing,fu2025icrt,tsagkas2026afa} over pretrained visual
features
to learn ROIs from demonstration actions.
Its attention is trained without spatial annotations or fixed rules
that convert trajectory events into crop locations.
Starting from a task- and state-conditioned query, it repeatedly
reads visual patches and updates the query with retrieved context,
allowing spatial focus to adapt to the current observation.
Query-dependent head gating lets different spatial cues contribute
as robot state and task stage change.
The final readout yields a visual context for action prediction and
an attention map for ROI extraction.

\subsection{Seeker Architecture}
\label{sec:seeker_architecture}
\paragraph{Robot-conditioned Query.}
Before reading the image, Seeker initializes a task- and
state-conditioned query from task embedding $Z\in\mathbb{R}^{D}$
and proprioceptive state
$s=[s_{\mathrm{eef}},s_{\mathrm{grip}}]\in\mathbb{R}^{S}$:
\begin{equation*}
Q =\mathrm{FiLM}(Z,s)
=\left(1+\gamma(s)\right)\odot Z+\beta(s),
\end{equation*}
where $\gamma$ and $\beta$ are predicted by an MLP, and
$\odot$ denotes element-wise multiplication.

\paragraph{Iterative Gated Readout.}
For a $224\times224$ image, frozen DINOv3~\cite{dinov3_2025}
produces a $14\times14$ grid of patch features
$x\in\mathbb{R}^{N_p\times D}$, with $N_p=196$.
Linear projections form keys and values
$K,V\in\mathbb{R}^{N_p\times D}$.
At each of $T$ lookups, the query attends to these features:
\begin{equation*}
Q_*=\phi_q(Q),
\qquad
\{A_h,C_h\}_{h=1}^{H}=\mathrm{MHA}(Q_*,K,V),
\end{equation*}
where $\phi_q$ is a learned projection, $\mathrm{MHA}$ denotes
multi-head attention, and $A_h$ and $C_h$ are the attention map
and context from head $h$.
To account for head specialization, Seeker assigns each head
a query-dependent weight:
\begin{equation*}
\omega_h
=
\mathrm{softmax}_{h}\!\left(
    \left\langle C_h,\phi_g(\bar Q_*)\right\rangle
\right).
\end{equation*}
Here $\bar Q_*$ is the mean of the head-wise query components,
and $\phi_g$ is a learned projection.
The gated context and attention map are
\begin{equation*}
\widetilde C=\sum_{h=1}^{H}\omega_h C_h,
\qquad
\widetilde A=\sum_{h=1}^{H}\omega_h A_h.
\end{equation*}
For the first $T-1$ lookups, the retrieved context updates
the query:
\begin{equation*}
Q\leftarrow\mathrm{FiLM}(Q,\widetilde C).
\end{equation*}
At the final lookup, $\widetilde C$ is passed with task embeddings and robot states 
to the training-only diffusion action head, while
$\widetilde A$ forms the ROI map.
We use $T=3$ and ablate the number of lookups in
Sec.~\ref{sec:ablation}.
Head-gating visualizations in
App.~\ref{app:training:head_gating}
show how the weights vary with robot state and task stage.

\paragraph{Action-supervised ROI Interface.}
\label{subsec:seeker_training}
We train Seeker with a temporary diffusion action head using the
standard noise-prediction objective~\cite{chi2023diffusionpolicy}.
Because action supervision alone can yield spatially broad attention
maps, we use staged coarse-to-fine training to obtain compact ROIs.
We first train a coarse Seeker, then introduce a separate fine Seeker
with its own diffusion action head, operating only on the shared
DINO tokens within the coarse ROI.
The fine Seeker's predicted box guides attention trimming in the
coarse Seeker, suppressing attention outside this box.
After training, we discard the fine Seeker and both action heads and freeze
the coarse Seeker for ROI prediction.
Further details on the mechanism, training schedule, and trimming objective are provided in App.~\ref{app:seeker_implementation_details}.

For ROI extraction, we rank patches by the final attention $\widetilde A$ and retain
the fewest whose cumulative mass exceeds \texttt{top\_p}
(Fig.~\ref{fig:seeker_overview}(b)).
The selected patches form a mask, and their enclosing box defines
the ROI crop.
In practice, we set the threshold to 0.8, retaining most of the attention mass while removing the low-weight tail for compact ROI extraction.

The mask and box support three downstream uses.
\textbf{(1) RGB cropping}: the box crops and resizes the static third-person
view. We FiLM-condition policy features on its location and scale,
while leaving eye-in-hand views uncropped.
\textbf{(2) Point-cloud filtering}: the box filters points by their
image-plane projections before downsampling.
\textbf{(3) Mask-guided augmentation}: the mask preserves selected image
regions while appearance outside them is perturbed.
ROIs are precomputed for downstream policy training, while frozen
Seeker extracts them from new observations at inference.

\section{Simulation Evaluations}
\label{sec:simulation}

\begin{figure*}[t]
\centering

\subfloat[Task Overview]{
    \includegraphics[height=0.22\textwidth]{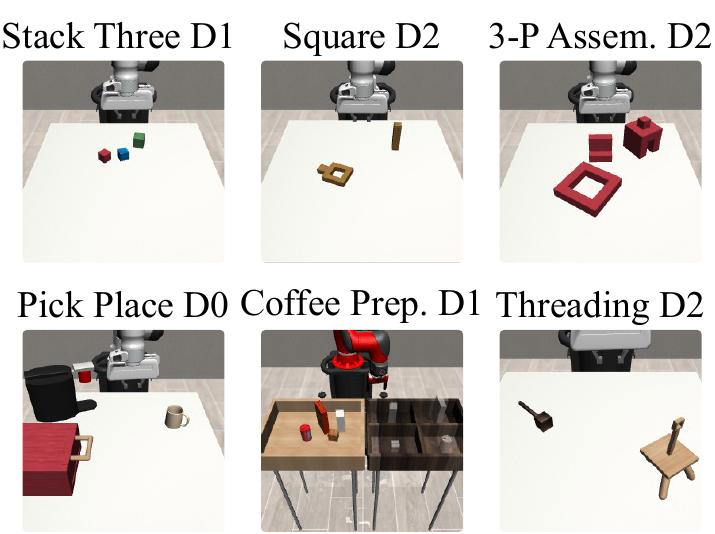}
    \label{fig:seeker_vis_task}
}
\hfill
\subfloat[Progress-aware Seeker ROIs]{
    \includegraphics[height=0.20\textwidth]{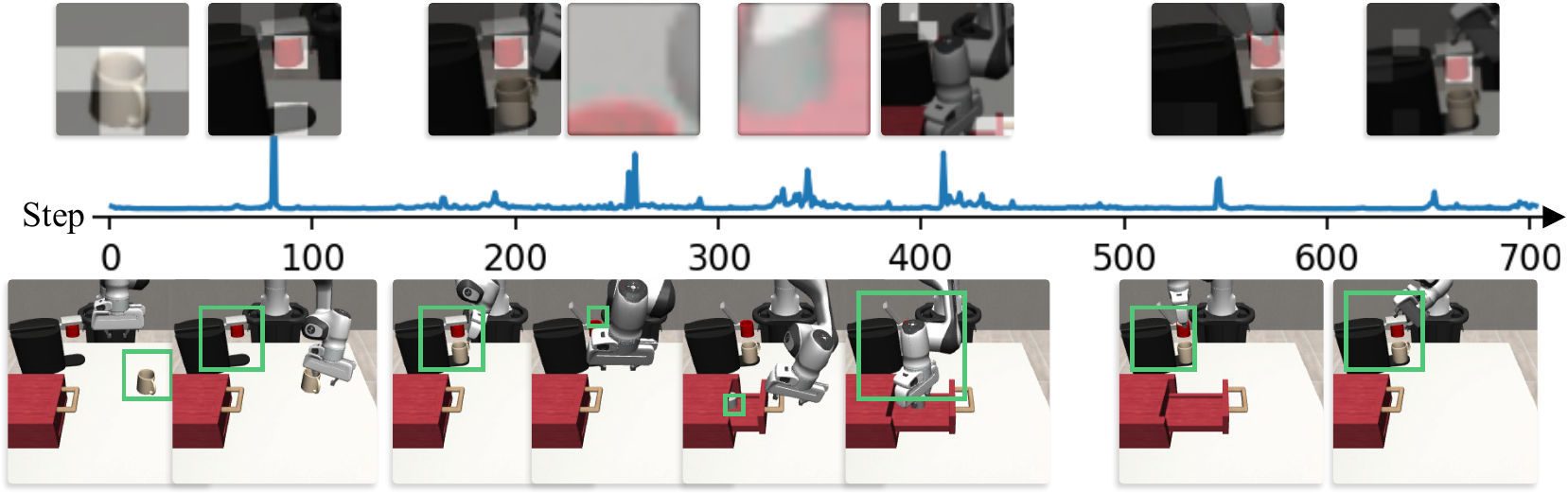}
    \label{fig:seeker_vis_temporal}
}
\\
\vspace{-1mm}
\subfloat[Heuristic ROI Stress Cases]{
    \includegraphics[height=0.15\textwidth]{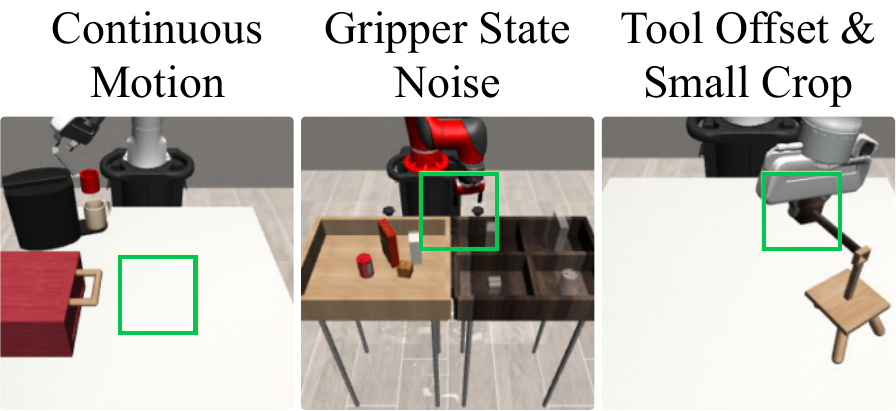}
    \label{fig:seeker_vis_rvt2_failure}
}
\hfill
\subfloat[Seeker Crops and Masks]{
    \includegraphics[height=0.15\textwidth]{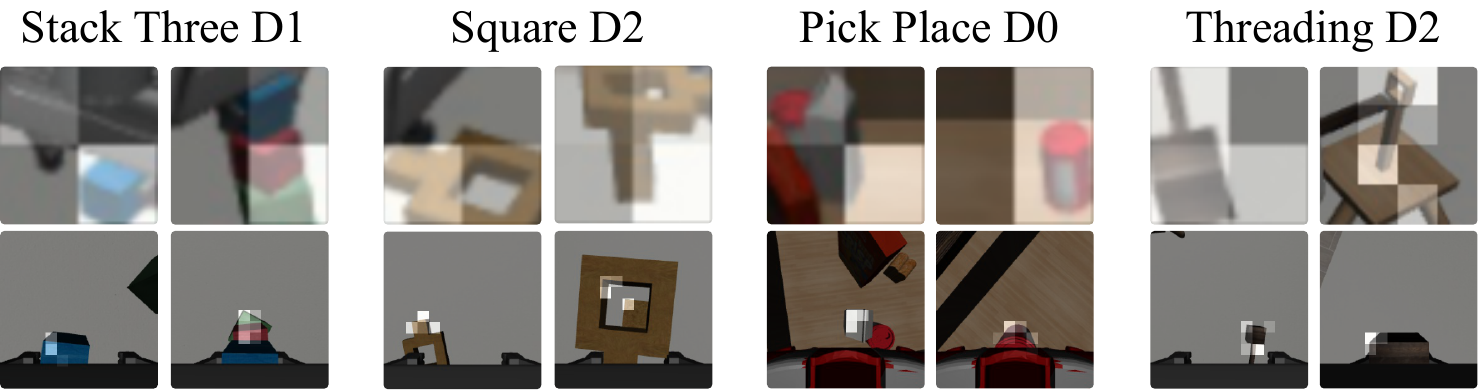}
    \label{fig:seeker_ROI_and_mask}
}
\caption{
\textbf{Simulation setup and Seeker visualization.}
(a) MimicGen~\cite{mandlekar2023mimicgen} task overview.
(b) In Coffee Preparation, Seeker ROIs shift with task progress; selected frames correspond to peaks in pairwise context-feature distance.
(c) RVT2-style heuristic ROI stress cases: missed continuous-motion cues, noisy gripper-triggered keyframes, and tool-contact regions offset from the TCP proxy.
(d) Seeker crops and masks across tasks.
}
\label{fig:seeker_vis}
\vspace{-2mm}
\end{figure*}

\paragraph{Experimental Protocol.}
\label{sec:sim_protocol}
Fig.~\ref{fig:seeker_vis_task} illustrates the six MimicGen~\cite{mandlekar2023mimicgen} tasks used in our simulation study: Stack Three D1, Square D2, Three-Piece Assembly D2, Coffee Preparation D1, Pick \& Place D0, and Threading D2.
These tasks cover high spatial variance and long-horizon manipulation; detailed task descriptions are provided in App.~\ref{app:task_setup}.
We omit D0/D1/D2 suffixes where clear.

We train a \textbf{single} multi-task Seeker once, using 100 demonstrations per task from the original-background RGB data, and keep it frozen for all simulation experiments.
This design isolates the ROI interface from downstream policy learning: Seeker remains unchanged even when downstream policies are trained with 1,000 demonstrations per task, randomized backgrounds, or point-cloud inputs.
We evaluate this fixed ROI interface through RGB data efficiency, background generalization, and point-cloud transfer, testing whether action supervision yields a reusable spatial bottleneck across input protocols, appearance shifts, and sensing modalities.

\paragraph{RGB Data Efficiency.}
\label{sec:sim_exp_multiview}

The RGB comparison isolates Seeker's action-grounded visual bottleneck under a shared downstream stack.
All controlled baselines use the same pretrained ResNet-18 policy encoder, since pretrained features substantially improve RGB manipulation performance~\cite{klee2026raven}.
In addition to vanilla Diffusion Policy~\cite{chi2023diffusionpolicy}, MirrorAug~\cite{mirrorduo2025} provides an augmented no-crop reference, applying reflection-consistent image--action augmentation.
All controlled RGB policies use $84\times84$ CNN inputs in both views; Seeker localizes at $224\times224$ and resizes the third-person ROI to the same policy input resolution.

RVT2-Crop is our RVT-2-inspired~\cite{goyal2024rvt2} crop baseline, predicting a 2D crop target from frozen DINOv3 features and proprioception.
Its heatmap supervision follows the original keypoint extraction recipe based on gripper-state changes and low end-effector velocity, testing whether this action-derived prior is sufficient for policy cropping.
Oracle ROI uses privileged stage-aware affordance pixels to crop target objects; it is not a control upper bound because contacts or inter-object relations may lie outside the crop.
RAVEN~\cite{klee2026raven} is included as an external equivariant-policy reference outside the controlled stack.

\begin{table}[t]
    \centering
    \centering
\footnotesize

\providecommand{\refnum}[1]{\cellcolor{black!4}{#1}}

\setlength{\tabcolsep}{3.6pt}

\begin{tabular}{l *{7}{c}}
\toprule
&
\textbf{Stack Three} &
\textbf{Square} &
\textbf{3-P Assembly} &
\textbf{Coffee Prep.} &
\textbf{Pick \& Place}$^\dagger$ &
\textbf{Threading} &
\textbf{Avg.} \\
\midrule

& \multicolumn{7}{c}{\textit{Input-level Comparison}} \\
\cmidrule(lr){2-8}

DiffPo (Pre) &
52.7 & 22.0 & 14.0 & 65.3 & 20.0 & \refnum{20.7} & 32.5 \\

MirrorAug &
68.0 & \refnum{32.7} & 20.7 & 62.0 & 20.7 & 18.7 & 37.1 \\

RVT2-Crop &
\refnum{71.3} & 26.0 & \refnum{26.7} & \refnum{69.3} &
\refnum{49.3} & 12.7 & \refnum{42.6} \\

\addlinespace[0.5pt]

\textbf{Seeker (ours)} &
\textbf{81.3} & \textbf{46.0} & \textbf{58.7} &
\textbf{82.0} & \textbf{69.3} & \textbf{38.0} & \textbf{62.6} \\

\textit{Gain} (rel., \%) &
\gainup{14.0} & \gainup{40.7} & \gainup{119.9} &
\gainup{18.3} & \gainup{40.6} & \gainup{83.6} & \gainup{52.8} \\

\midrule

& \multicolumn{7}{c}{\textit{External SOTA and Privileged ROI References}} \\
\cmidrule(lr){2-8}

RAVEN &
80.7 & 50.0 & 26.7 & 72.7 & 54.7 & 28.0 & 52.1 \\

Oracle ROI$^\ast$ &
80.7 & 44.7 & 62.7 & 81.3 & 71.3 & 44.7 & 64.2 \\

\bottomrule
\end{tabular}

    \vspace{1mm}
    \caption{
    \textbf{Data Efficiency on MimicGen with 100 Demos.}
    Success rate (\%) averaged over three seeds; standard deviations in Tab.~\ref{app:tab:main_sim_std}.
    Gray cells mark the strongest non-Seeker input-level baseline.
    $^\ast$The oracle provides stage-aware affordance ROIs.
    $^\dagger$Pick \& Place uses strict episode success with all baselines rerun.
    }
    \label{tab:mimicgen_data_efficiency}
    \vspace{-4mm}
\end{table}

Tab.~\ref{tab:mimicgen_data_efficiency} shows that Seeker improves average success from $42.6\%$ for the strongest non-Seeker input-level baseline to $62.6\%$, with a mean task-wise relative gain of $52.8\%$ over the strongest baseline on each task.
Seeker also exceeds the external SOTA reference RAVEN on average ($62.6$ vs.~$52.1\%$) and nearly matches Oracle ROI ($64.2\%$), despite using no scripted affordance boxes, spatial labels, gaze, or language grounding.

RVT2-Crop is the strongest controlled baseline on average, but its uneven gains highlight limitations of fixed keyframe and proxy assumptions: these can miss continuous-motion cues, produce noisy gripper-triggered targets, or focus on TCP proxies offset from the interaction region (Fig.~\ref{fig:seeker_vis_rvt2_failure}).
Seeker improves consistently across tasks, including spatially ambiguous or contact-rich settings where policies must localize small control-relevant regions across objects, tools, and task stages.
Increasing the downstream training budget tenfold, from 100 to 1,000 demonstrations per task, still does not close the gap: Seeker retains an advantage over MirrorAug while reusing the same frozen ROI predictor trained with only 100 demonstrations per task (App.~\ref{app:experiments:high_data}). This suggests that even substantially more demonstration data cannot fully compensate for the lack of an explicit spatial bottleneck.

We also evaluate policies using DINOv3 features directly, either through the global CLS token or Seeker's attention-pooled context without ROI extraction.
Both fall short of the crop-based Seeker policy and can even underperform the pretrained ResNet-18 Diffusion Policy baseline on individual tasks, suggesting that access to DINO features alone does not explain Seeker's gains (see Sec.~\ref{sec:ablation}).

\paragraph{Mask-Guided Augmentation for Background Generalization.}
\label{sec:background_generalization}
\label{sec:background_baselines}

Beyond data efficiency, we reuse the same frozen Seeker trained on the original training set under background shifts.
Its mask guides image superimposition, preserving predicted control-critical regions more strongly while applying larger appearance perturbations elsewhere.
Given an image $I$ and texture image $I_{\mathrm{tex}}$, we form
\begin{equation*}
I_{\mathrm{aug}}
=
M\odot I+(1-M)\odot I_{\mathrm{tex}},
\end{equation*}
where $M\in[0,1]^{H\times W}$ is the spatial preservation mask,
broadcast over color channels, and $\odot$ denotes element-wise multiplication.
Random Overlay~\cite{mirrorduo2025} uses a spatially constant preservation weight.
Guided variants instead use spatially varying weights: RVT2-Crop uses its Gaussian heatmap for the front view.
For Seeker, we upsample and max-normalize the predicted attention mask, then clamp it to $[0.3,0.8]$ before blending.
This retains some original visual evidence throughout the image while perturbing high-attention regions less, accounting for the mask's coarse patch-level resolution.

We evaluate Stack Three D1, Square D2, and Three-Piece Assembly D2 under shuffled tabletop textures.
Policies are trained either on the original $100$ demonstrations or on the same trajectories re-rendered with $25$ shuffled tabletop backgrounds.
Augmentation examples and implementation details are shown in Fig.~\ref{fig:background_aug_and_eval} and App.~\ref{app:experiments:guided_aug}.

\begin{figure}[t]
    \input{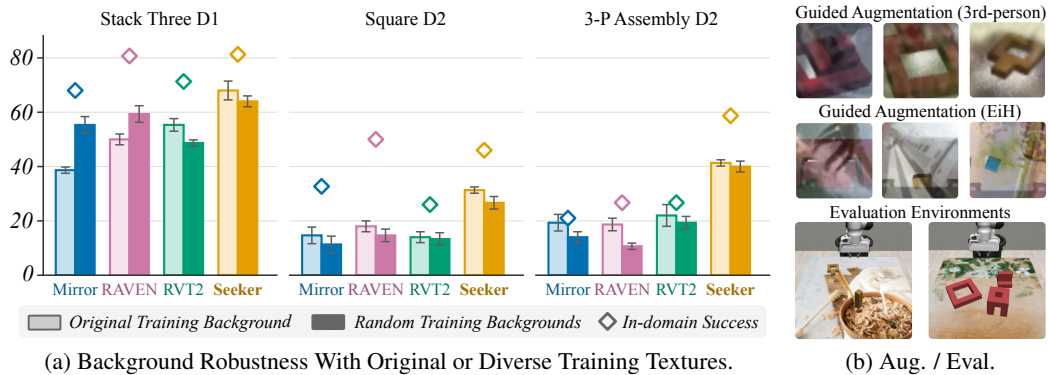}
    \vspace{-1mm}
    \caption{
    \textbf{Background Robustness and Guided Augmentation.}
    (a) Success rates (\%). Light bars use original-background training, solid bars use training with 25 randomized backgrounds, and diamonds show in-domain mean success from the main benchmark.
    We compare MirrorAug, RAVEN+Overlay, RVT2-Crop, and Seeker-Guided augmentation.
    (b) Seeker-guided augmentation examples and evaluation environments.
    }
    \label{fig:background_shuffling}
    \vspace{-2mm}
\end{figure}

Fig.~\ref{fig:background_robustness} shows that Seeker-Guided achieves the highest average OOD success with both original- and randomized-background training.
This indicates that the action-grounded regions identified by the frozen Seeker remain useful under background changes and can guide appearance augmentation.
With only 100 demonstrations, complex textures can make task-relevant cues harder to learn, outweighing the benefit of background diversity.
OOD performance in this regime therefore conflates in-domain learning and robustness to shift.
At 1,000 demonstrations per task, stronger in-domain performance reduces this confound: absolute success and relative drop agree, with Seeker retaining its advantage on both metrics across all three tasks.
Background randomization also consistently improves methods' performance at this larger budget (App.~\ref{app:experiments:high_data}, Tab.~\ref{tab:sim_high_data_ood}).

\paragraph{Point-cloud Transfer.}
\label{sec:exp:pointcloud}

\begin{figure}[h]
    \centering

\begin{minipage}[c]{0.56\linewidth}
\centering
\includegraphics[width=\linewidth]{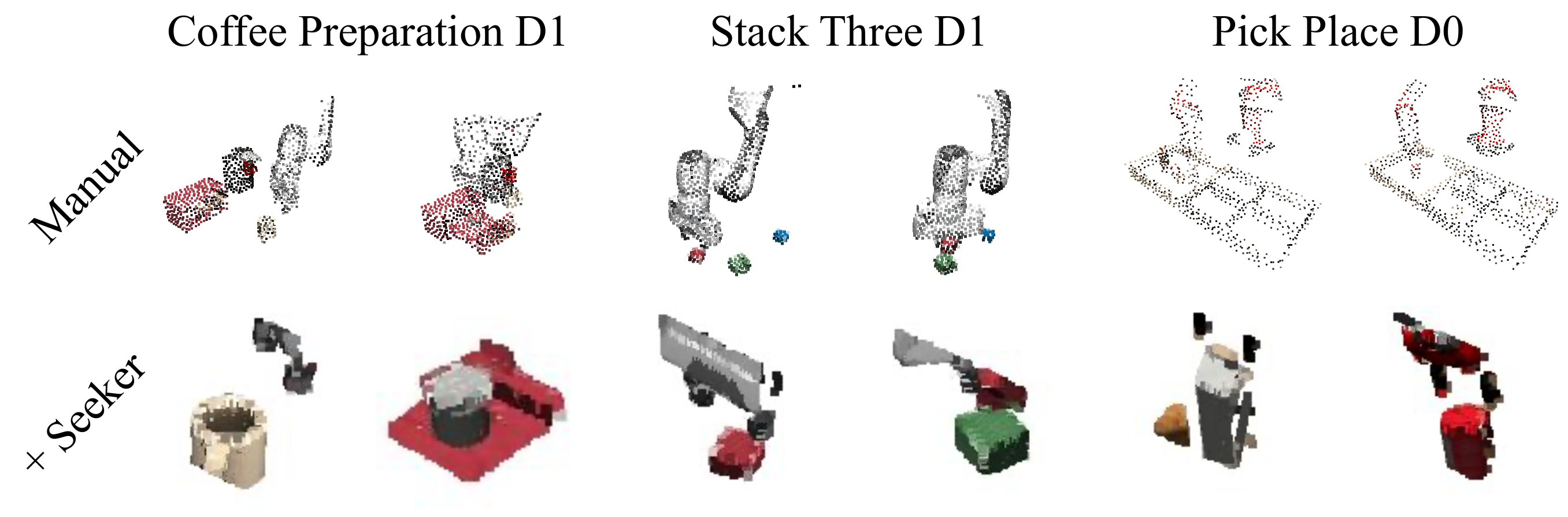}
\end{minipage}
\hfill
\begin{minipage}[c]{0.42\linewidth}
\centering
\scriptsize
\setlength{\tabcolsep}{2.7pt}
\renewcommand{\arraystretch}{1.08}

\begin{tabular}{@{}ccccc@{}}
    \toprule
    \textit{Manual} &
    \textit{Seeker} &
    \textbf{Pick Place} &
    \textbf{Stack-3} &
    \textbf{Coffee Prep.} \\
    \midrule
    \xmark & \xmark &
    $0.0$ & $0.0$ & $0.0$ \\

    \cmark & \xmark &
    \textit{1.3} & \textit{23.3} & \textit{22.0} \\

    \cmark & \cmark &
    $\mathbf{26.0}$ & $\mathbf{47.3}$ & $\mathbf{64.3}$ \\

    \addlinespace[0.5pt]
    \multicolumn{2}{c}{\textit{Gain} (abs.)} &
    \gainup{24.7} & \gainup{24.0} & \gainup{42.3} \\
    \bottomrule
\end{tabular}

\vspace{0.5mm}
{\footnotesize
\textbf{DP3 point-cloud transfer.}
}

\end{minipage}

    \caption{
    \textbf{Point-cloud transfer with DP3~\cite{ze20243ddif}.}
    Left: manual workspace cropping vs. manual+Seeker ROI filtering, with point clouds independently scaled for visualization.
    Right: success rates with 200 demonstrations for no crop, manual crop, and manual+Seeker filtering; absolute gains are computed over manual cropping.
    }
    \label{fig:pointcloud_transfer}
    \vspace{-1mm}
\end{figure}

We evaluate whether Seeker's image-plane ROI can serve as a plug-and-play spatial filter for point-cloud policies.
Seeker is not retrained for point clouds; we reuse the same frozen model trained on RGB observations and actions.
Given a predicted Seeker box, we remove points whose image-plane projections fall outside the ROI, then apply farthest-point sampling to match the fixed input size required by DP3~\cite{ze20243ddif}.
Since point-cloud policy learning commonly relies on manually cropping a task-specific workspace to remove background and tabletop regions~\cite{chisari2024learning, ze20243ddif, wang2024equivariant}, we evaluate Seeker as a complementary ROI filter on top of this manual workspace crop.
This transfer study uses 200 demonstrations because, with 100 demonstrations, the manual-crop baseline was too weak to support informative relative comparisons.
As shown in Fig.~\ref{fig:pointcloud_transfer}, adding Seeker ROI filtering makes the retained point clouds more task-focused and improves DP3 over manual workspace cropping by 30.3 absolute points on average on the three tasks, despite Seeker being trained only from RGB action supervision.
Overall, DP3 remains below image-based policies (Tab.~\ref{tab:mimicgen_data_efficiency}), consistent with prior work~\cite{wang2024equivariant}.
Seeker-only filtering and full per-task analyses are provided in App.~\ref{app:experiments:pointcloud}.

\section{Real-world Evaluations}
\label{sec:real_world_evaluation}

\begin{figure*}[t]
    \input{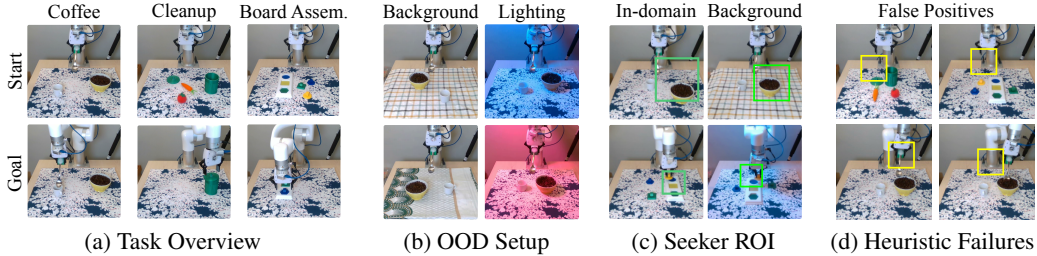}
    \vspace{-1mm}
    \caption{
    \textbf{Real-world Setup and Visual Diagnostics.}
    (a) Start and goal configurations for the three real-world tasks.
    (b) OOD evaluation conditions with background and lighting shifts.
    (c) Seeker ROIs remain focused on predicted control-relevant regions under appearance changes.
    (d) Heuristic crops can produce ghost keyframes or miss interaction-relevant continuous-motion cues without reliable gripper-state anchors.
    }
    \label{fig:real_exp_viz}
    \vspace{-2mm}
\end{figure*}

\paragraph{Experimental Setup.}
We evaluate Seeker and two representative baselines on a real-world multi-view RGB setup with a UFactory xArm7, using a fixed third-person camera and an eye-in-hand camera.
Seeker localizes ROIs from $224\times224$ images and resizes the third-person crop to $84\times84$ for the policy.
All downstream policy CNNs receive $84\times84$ inputs in both views, with the eye-in-hand view left uncropped.
Seeker policies are trained with mask-guided augmentation, and the same policy checkpoint is evaluated under ID and OOD conditions.

The three tasks are performed on visually distracting table textures and require different forms of spatial attention (Fig.~\ref{fig:real_exp:task_overview}):
\textit{Coffee Transport} is a low-tolerance continuous manipulation task, where the robot must carry coffee beans on a spoon and deposit them into a small white espresso cup under low foreground--background contrast;
\textit{Table Cleanup} is a long-horizon sequential insertion task requiring 3D object rotations, including tilting a carrot so that it fits through the bin opening; and
\textit{Board Assembly} requires precise shape-based placement despite color mismatch between the objects and target slots.
Detailed real-world task descriptions are provided in App.~\ref{app:real_world_exp}.

\begin{table}[t]
    \centering
\footnotesize

\providecommand{\refnum}[1]{\cellcolor{black!4}{#1}}

\setlength{\tabcolsep}{4.5pt}
\renewcommand{\arraystretch}{0.96}

\begin{tabular}{l *{12}{c}}
\toprule
\multirow{2}{*}{\textbf{Method}}
& \multicolumn{3}{c}{\textbf{Coffee Trans.}}
& \multicolumn{3}{c}{\textbf{Table Cleanup}}
& \multicolumn{3}{c}{\textbf{Board Assem.}}
& \multicolumn{3}{c}{\textbf{Average}} \\
\cmidrule(lr){2-4} \cmidrule(lr){5-7} \cmidrule(lr){8-10} \cmidrule(lr){11-13}
& \textit{ID} & \textit{Light.} & \textit{Bg.}
& \textit{ID} & \textit{Light.} & \textit{Bg.}
& \textit{ID} & \textit{Light.} & \textit{Bg.}
& \textit{ID} & \textit{OOD} & \textit{Retention} \\
\midrule

MirrorAug
& \refnum{45} & \refnum{30} & 0
& 40 & 10 & 10
& 20 & 0 & 20
& 35.0 & 11.7 & 33.3 \\

RVT2-Crop
& 40 & 15 & 0
& \refnum{60} & \refnum{25} & \refnum{35}
& \refnum{45} & \refnum{20} & \refnum{25}
& \refnum{48.3} & \refnum{20.0} & \refnum{41.4} \\

\addlinespace[0.5pt]

\textbf{Seeker (ours)}
& \textbf{85} & \textbf{80} & \textbf{35}
& \textbf{80} & \textbf{75} & \textbf{60}
& \textbf{65} & \textbf{60} & \textbf{50}
& \textbf{76.7} & \textbf{60.0} & \textbf{78.3} \\

\bottomrule
\end{tabular}

    \vspace{1mm}
    \caption{
    \textbf{Real-world experiments.}
    Success rate (\%) over 20 rollouts under in-domain (\textit{ID}), lighting-shift (\textit{Light.}), and background-shift (\textit{Bg.}) evaluations.
    Gray cells mark the strongest non-Seeker baseline in each column.
    \textit{Avg. OOD} averages lighting and background shifts; \textit{Ret.} is the retention ratio, \textit{Avg. OOD}/\textit{Avg. ID}.
    }
    \label{tab:real_exp_results}
    \vspace{-4mm}
\end{table}

\paragraph{Policy Success and Generalization.}
Tab.~\ref{tab:real_exp_results} shows that Seeker improves average in-domain success from the best baseline's $48.3$ to $76.7\%$ and raises average OOD success from $20.0$ to $60.0\%$.
Its $78.3\%$ retention indicates that the learned ROI remains usable under real appearance changes, including lighting shifts that overexpose the eye-in-hand view.

\paragraph{Heuristic Crop Failure Analysis.}
The RVT2-Crop results highlight two settings where action-heuristic crop supervision is challenged.
First, in Table Cleanup and Board Assembly, inconsistent velocity profiles across demonstrations can create false-positive low-speed keyframes, causing the crop predictor to attend to spurious targets before reaching the relevant interaction region.
Second, in Coffee Transport, continuous spoon motion provides no reliable gripper-state anchor, so the heuristic crop can miss the cup--spoon interaction and lock onto rotation-only or final end-effector poses (Fig.~\ref{fig:real_exp:rvt2_failures}).
When this crop is misaligned, the policy loses useful third-person context and must rely more on the eye-in-hand stream, which is fragile when the white cup leaves the wrist-camera field of view or is washed out by lighting.

\section{Ablations}
\label{sec:ablation}
\begin{table}[!ht]
\centering
\footnotesize
\setlength{\tabcolsep}{3.6pt}
\begin{tabular}{l *{7}{c}}
\toprule
&
\textbf{Stack Three} &
\textbf{Square} &
\textbf{3-P Assembly} &
\textbf{Coffee Prep.} &
\textbf{Pick \& Place} &
\textbf{Threading} &
\textbf{Avg.} \\
\midrule

& \multicolumn{7}{c}{\textit{Policy Representation}} \\
\cmidrule(lr){2-8}

DINO CLS &
62.7 & 30.0 & 29.3 & 60.0 & 27.3 & 25.3 & 39.1 \\

Seeker Readout &
78.0 & 28.7 & 30.7 & 62.0 & 28.7 & 21.3 & 41.6 \\

\midrule

& \multicolumn{7}{c}{\textit{ROI Interface}} \\
\cmidrule(lr){2-8}

FiLM-only &
64.7 & 29.3 & 16.7 & 69.3 & 28.0 & 28.7 & 39.4 \\

Low-res Crop &
82.7 & 46.7 & 56.7 & 75.3 & 70.0 & 36.0 & 61.2 \\

\midrule

\textbf{Seeker (ours)} &
\textbf{81.3} & \textbf{46.0} & \textbf{58.7} &
\textbf{82.0} & \textbf{69.3} & \textbf{38.0} & \textbf{62.6} \\

\bottomrule
\end{tabular}
\vspace{1mm}
\caption{
\textbf{Policy Representation and ROI Interface Ablations (100 demos).}
Success rate (\%) averaged over three seeds.
The two groups test policy representations and ROI interfaces.
}
\label{tab:seeker_ablations}
\vspace{-4mm}
\end{table}

\begin{figure}[!ht]
    \centering
    \begin{minipage}[t]{0.56\linewidth}
        \vspace{0pt}
        \begin{minipage}[c][2.8cm][c]{\linewidth}
            \centering
            \includegraphics[
                height=2.8cm,
                width=\linewidth,
                keepaspectratio
            ]{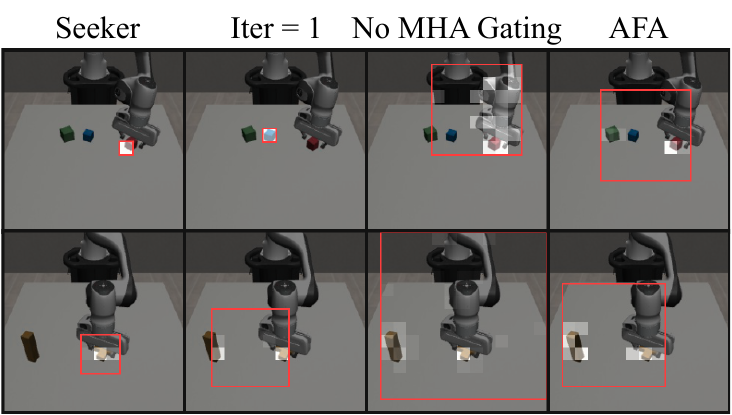}
        \end{minipage}
        \caption{
        \textbf{Readout Diagnostics.}
        ROIs on Stack Three D1 (top) and Square D2 (bottom):
        default Seeker, a single lookup, uniform head
        averaging, and AFA-Crop.
        }
        \label{fig:seeker_refinement}
    \end{minipage}
    \hfill
    \begin{minipage}[t]{0.40\linewidth}
        \captionsetup{type=table}
        \vspace{0pt}
        \begin{minipage}[c][2.8cm][c]{\linewidth}
            \centering
            \footnotesize
            \setlength{\tabcolsep}{5pt}
            \renewcommand{\arraystretch}{1.1}
            \begin{tabular}{@{}lcc@{}}
                \toprule
                Method & \textbf{Stack-3} & \textbf{Square} \\
                \midrule
                AFA-Crop & 55.3 & 26.0 \\
                Seeker, $T=1$ & 71.3 & 32.7 \\
                Seeker, $T=3$ & 81.3 & 46.0 \\
                \bottomrule
            \end{tabular}
        \end{minipage}
        \caption{
        \textbf{Readout and Refinement.}
        Success (\%) with 100 demonstrations,
        averaged over three seeds.
        }
        \label{tab:seeker_refinement}
    \end{minipage}
    \vspace{-3mm}
\end{figure}

\paragraph{Visual Representation, Cropping, and Resolution.}
\label{sec:ablation_resolution_and_crop}
We first test whether Seeker's gains arise from DINO features,
box conditioning, or explicit cropping.
We fine-tune DINOv3 as the third-person policy encoder using
either its global CLS token or Seeker's attention-pooled context
without ROI extraction.
These variants achieve $39.1\%$ and $41.6\%$ average success,
respectively, compared with $62.6\%$ for full Seeker
(Tab.~\ref{tab:seeker_ablations}).
Neither the pretrained features nor the learned context alone
recovers the crop-based policy's performance.

FiLM-only conditions full-image CNN features on the predicted
box location and scale while retaining the ROI predictor.
It is FLOP-matched to Seeker but substantially weaker, indicating
that additional computation and box conditioning alone do not
explain the gain.
Low-res Crop applies the same normalized box to the downsampled
policy-resolution image and nearly matches Seeker
($61.2$ vs.~$62.6\%$).
This suggests that the main benefit comes from explicit spatial
cropping, with higher-resolution ROI content contributing a
smaller additional gain.

\paragraph{Attention Readout and Iterative Refinement.}
\label{sec:ablation_refinement}
We compare Seeker with a cropping baseline derived from AFA~\cite{tsagkas2026afa} (AFA-Crop), which trains
a single learned-query cross-attention readout over frozen visual
features through action supervision.
We convert its attention map into an ROI crop for the downstream policy.
Tab.~\ref{tab:seeker_refinement} shows that AFA-Crop falls below
Seeker on both tasks, while increasing Seeker from one to three
lookups improves success by $10.0$ points on Stack Three and
$13.3$ points on Square.
With $T=1$, attention depends on the current image, but the query
has not incorporated retrieved visual evidence.
Fig.~\ref{fig:seeker_refinement} illustrates how further lookups
sharpen or correct localization, while uniform head averaging
can produce diffuse ROIs or emphasize irrelevant robot regions.

\paragraph{Training Objective.}
\label{sec:ablation_objectives}
We compare diffusion-based Seeker training with direct regression,
flow matching~\cite{lipman_2023_flow}, and IMLE~\cite{rana2025imle}.
Direct regression yields noisy attention, IMLE cleaner but diffuse maps,
and flow matching tighter masks closest to diffusion.
Mean ROI box IoUs with diffusion-trained Seeker are $0.30$, $0.50$,
and $0.67$ for direct regression, IMLE, and flow matching, respectively.
These measure agreement with diffusion rather than ground-truth
localization accuracy. App.~\ref{app:training:objective_ablation}
details the IoU evaluation protocol and provides qualitative comparisons
in Fig.~\ref{fig:seeker_training_objectives}.

\section{Limitations}

\paragraph{Single-task Data Efficiency and Multi-task Data Overhead.}
Seeker can be pretrained on fewer demonstrations than the downstream policy when they contain sufficient spatial variation in control-relevant interaction regions and task phases.
App.~\ref{app:spatial_coverage} shows that a spatially diverse subset equivalent to 25 demonstrations suffices for single-task ROI learning.
However, 25 demonstrations per task are insufficient for multi-task co-training in our experiments.
We therefore use 100 demonstrations per task, while leaving the minimum sufficient budget between 25 and 100 undetermined.

\paragraph{Weak Visual Supervision and Attention Priors.}
In spatially invariant or visually redundant phases, actions may be predictable from proprioception or trajectory regularities, giving little incentive to localize visual evidence.
This has limited in-domain impact in Tab.~\ref{tab:mimicgen_data_efficiency}, where Seeker matches or approaches Oracle ROI on Coffee Preparation and Pick \& Place despite largely static targets.
Generalization to layouts that vary these targets may require additional spatially varied demonstrations or retraining.
An attention prior can also bootstrap Seeker: preliminary experiments show that a temporary RVT2-style keypose heatmap used as an auxiliary attention target can recover localization.
When this prior is also uninformative, additional grounding cues may be needed.

\textbf{Training and Inference Overhead.}
Seeker adds a separate pretraining stage before the downstream policy, after which it is frozen and reused.
For consistency, our main experiments train Seeker on the full policy-training set, giving a conservative total training overhead of roughly $1.5$--$1.8{\times}$ over training the downstream policy alone.
This overhead can be reduced by training Seeker on fewer spatially diverse demonstrations, as suggested by App.~\ref{app:spatial_coverage}.
At inference time, ROI extraction adds approximately $12\,\mathrm{ms}$ per step in our implementation.

\section{Conclusion}
We presented Seeker, an action-supervised module that learns progression-aware ROIs from observation--action data without semantic boxes, gaze, language grounding, or affordance annotations.
By incorporating visual evidence into a task- and state-conditioned query, Seeker identifies where to attend as the task progresses, shifting part of the burden of scene understanding and task-progress inference away from the downstream policy so it can focus on local control.
Across simulation and real-robot experiments, this bottleneck improves data efficiency, supports targeted background augmentation, and transfers from RGB crops to point-cloud filtering.
Its advantage persists at larger downstream data budgets, suggesting that separating where to attend from how to act provides benefits that additional demonstrations alone do not fully recover.
These results point to action-grounded visual bottlenecks as a practical interface between perception and policy learning, and more broadly highlight that the visual structure recovered from action supervision depends on the coverage and variation present in the demonstrations.

\clearpage
\section*{Acknowledgements}
This work was partially supported by the Wallenberg AI, Autonomous Systems and Software Program (WASP), funded by the Knut and Alice Wallenberg Foundation, and by the Carl Zeiss Foundation through the ReScaLe project. Nick Heppert is additionally supported by the Konrad Zuse School of Excellence in Learning and Intelligent Systems (ELIZA) through the DAAD programme Konrad Zuse Schools of Excellence in Artificial Intelligence, sponsored by the Federal Ministry of Education and Research. Computational resources were provided on the Berzelius system funded by the Knut and Alice Wallenberg Foundation and operated by NAISS.
\bibliography{references}

\clearpage
\begin{center}
\vspace*{-4ex}

{\LARGE\bfseries
Attention from Action, for Action
\par}

\vspace{0.4ex}
\vspace{1.2ex}

{\large\bfseries APPENDIX}
\vspace{1.5ex}
\end{center}

This appendix provides implementation details, diagnostics, and extended results for Seeker.
App.~\ref{app:training} details Seeker training and ROI extraction.
App.~\ref{app:task_setup} summarizes the simulation tasks.
App.~\ref{app:experiments:high_data} reports RGB policy performance and background generalization with 1,000 demonstrations per task.
App.~\ref{app:spatial_coverage} studies data efficiency through spatial coverage.
App.~\ref{app:experimental_details} provides full simulation results and implementation details for RGB cropping, point-cloud filtering, and background augmentation.
App.~\ref{app:real_world_exp} describes the real-world tasks and training setup and provides additional qualitative visualizations.

\renewcommand{\thetable}{A.\Roman{table}}
\setcounter{table}{0}

\renewcommand{\thefigure}{A.\arabic{figure}}
\setcounter{figure}{0}

\renewcommand{\thesection}{\Alph{section}}
\renewcommand{\thesubsection}{\thesection.\arabic{subsection}}
\renewcommand{\thesubsubsection}{\thesubsection.\arabic{subsubsection}}
\renewcommand{\theHsection}{appendix.\Alph{section}}
\renewcommand{\theHsubsection}{appendix.\Alph{section}.\arabic{subsection}}
\renewcommand{\theHsubsubsection}{appendix.\Alph{section}.\arabic{subsection}.\arabic{subsubsection}}

\setcounter{section}{0}
\setcounter{equation}{0}
\setcounter{figure}{0}
\setcounter{table}{0}
\makeatletter
\renewcommand{\thesection}{\Alph{section}}
\renewcommand{\thesubsection}{\thesection.\arabic{subsection}}
\providecommand\thesubsectiondis{}
\renewcommand\thesubsectiondis{\arabic{subsection}.}
\renewcommand{\theHsection}{appendix.\Alph{section}}
\renewcommand{\theHsubsection}{appendix.\Alph{section}.\arabic{subsection}}
\renewcommand{\theHsubsubsection}{appendix.\Alph{section}.\arabic{subsection}.\arabic{subsubsection}}


\normalsize
\section{Seeker Training Details}
\label{app:training}

\subsection{Head Gating Visualizations}
\label{app:training:head_gating}

\begin{figure}[h]
    \centering
    \subfloat[Three-Piece Assembly\label{fig:head_gating_three_piece}]{
        \includegraphics[width=0.485\linewidth]{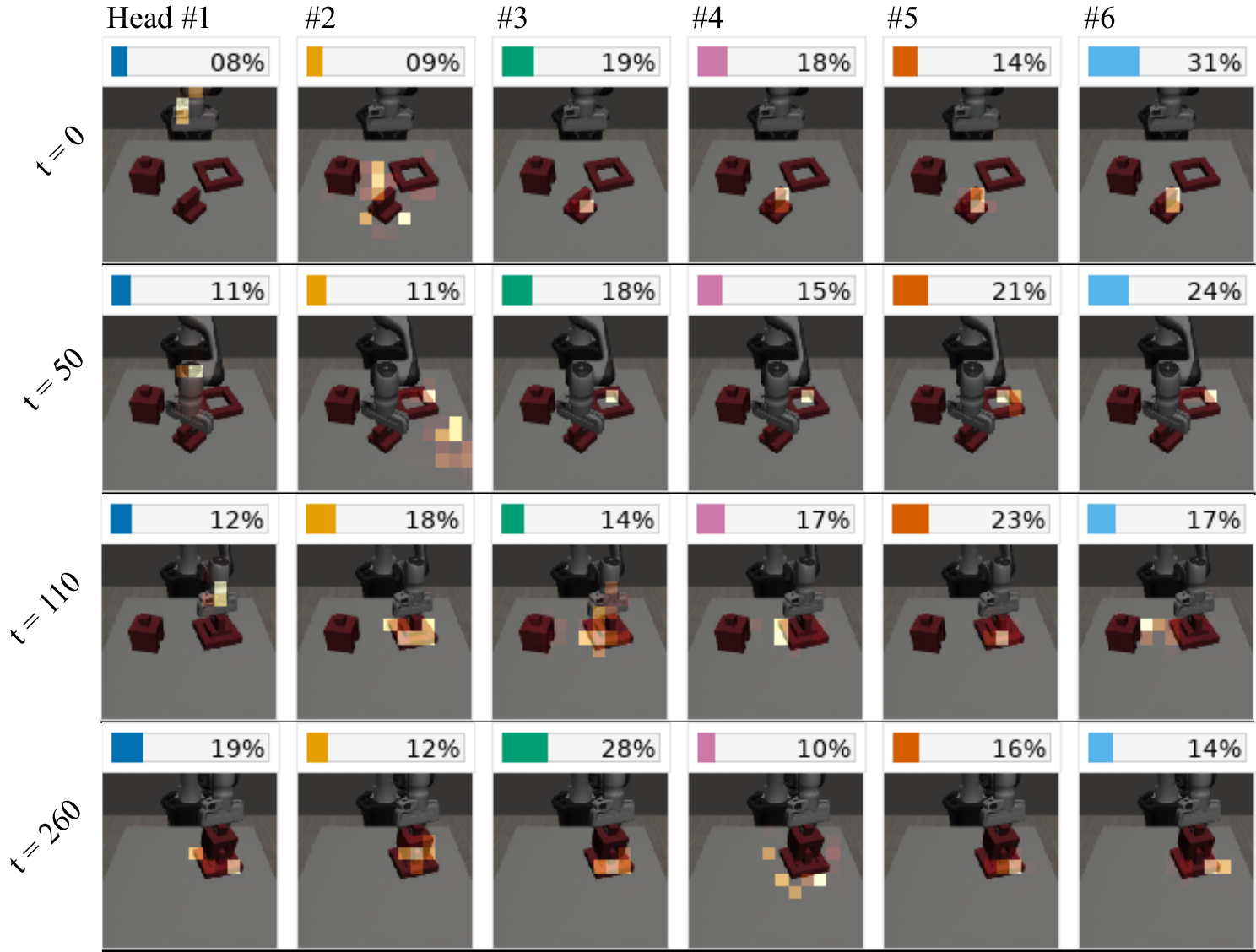}
    }
    \hfill
    \subfloat[Stack Three\label{fig:head_gating_stack_three}]{
        \includegraphics[width=0.485\linewidth]{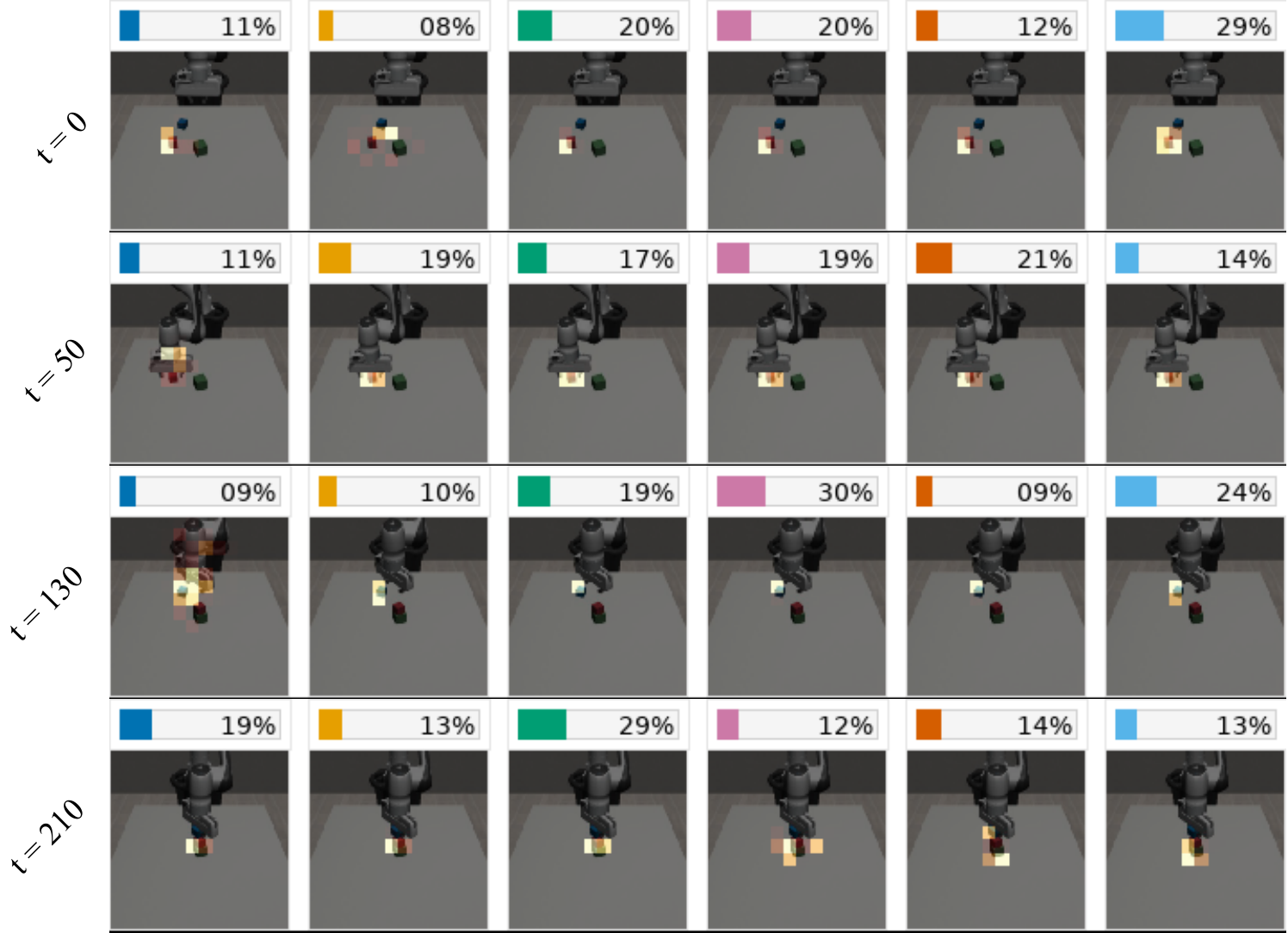}
    }
    \vspace{-1mm}
    \caption{\textbf{Per-frame head overlays and corresponding gating weights for two tasks.}}
    \label{fig:head_gating_vis}
    \vspace{-3mm}
\end{figure}

As shown in Fig.~\ref{fig:head_gating_vis}, the contribution of each attention head changes over the course of the task.
Taking the Three-Piece Assembly example, Head~1 remains relatively diffuse and becomes most salient for the final piece.
During the final placement stage, Head~4 becomes more diffuse, while Head~3 becomes the dominant contributor.

\subsection{Seeker Implementation Details}
\label{app:seeker_implementation_details}

\begin{figure}[h]
\centering
\includegraphics[width=0.55\linewidth]{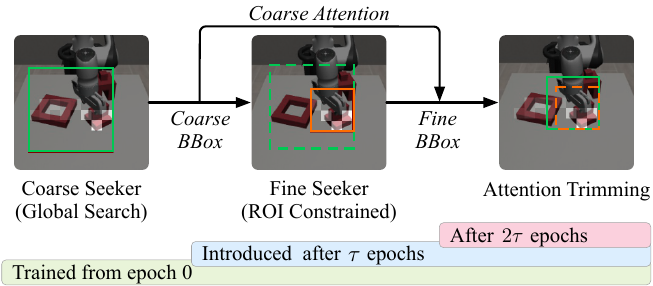}
\vspace{-1mm}
\caption{\textbf{Coarse-to-fine Attention Trimming.}
From epoch 0, Seeker produces a coarse box; after $\tau$ epochs, the fine branch predicts a fine box within the coarse ROI; after $2\tau$, attention trimming suppresses out-of-box mass and tightens the coarse box.
}
\label{fig:seeker_trimming}
\vspace{-4mm}
\end{figure}

\paragraph{Coarse-to-Fine Attention Trimming.}
\label{sec:attention_trimming}
Fig.~\ref{fig:seeker_trimming} shows that action supervision already steers Seeker attention toward task-relevant regions, but the resulting maps can remain heavy-tailed, yielding non-compact ROIs, especially in long-horizon tasks with shifting targets.
We adopt a coarse-to-fine staged training scheme (Fig.~\ref{fig:seeker_trimming}) with stage stride $\tau$ (epochs) to obtain compact, adaptive ROIs.
We train a \textbf{coarse} Seeker from the first epoch for localization, introduce a \textbf{fine} Seeker after $\tau$ epochs that operates only on patch tokens within the coarse ROI (box-cropped token set), and train both branches jointly with shared DINOv3 patch features but separate diffusion heads.
After $2\tau$ epochs, we activate attention trimming: letting {\small $\mathcal{M}^{\mathrm{fine}}_{\mathrm{box}}\in\{0,1\}^{N_p}$} be the binary patch mask induced by the fine ROI box (1 inside, 0 outside), we form a trimmed target attention for each head by masking and renormalizing the coarse last-step attention {\small $A_h^{(T)}$}, and minimize a head-weighted KL divergence,
\begin{equation*}
\scalebox{0.92}{$
\begin{array}{@{}c@{\qquad}c@{}}
\widehat{A}_h
=
\mathrm{stopgrad}\!\left[
\mathrm{Norm}\!\left(
\mathcal{M}^{\mathrm{fine}}_{\mathrm{box}} \odot A_h^{(T)}
\right)
\right],
&
\mathcal{L}_{\mathrm{trim}}
=
\sum_{h=1}^{H}\omega_h^{(T)}\,
D_{\mathrm{KL}}\!\left(\widehat{A}_h \,\|\, A_h^{(T)}\right).
\end{array}
$}
\end{equation*}
This suppresses residual out-of-box mass and tightens the coarse ROI until it largely overlaps the fine ROI.
After training, we discard the fine branch and retain only the coarse branch for single-pass ROI prediction at inference.

\paragraph{Iterative Refinement.}
Iterative refinement helps Seeker form task-progress-aware attention.
A single state-conditioned query can already localize useful regions because proprioception and gripper state provide strong phase cues, such as picking versus insertion in Square D2.
However, similar end-effector states can still correspond to different visual targets depending on object pose, scene layout, or task progress.
In such ambiguous cases, a single lookup can produce less compact attention, while refinement lets the query update from retrieved visual evidence before producing the final attention map.
In the default configuration, we use three lookups with two intervening query updates and derive the training attention map and pooled context from the final lookup.

\textbf{The low-dimensional inputs} consist of end-effector translation, end-effector rotation, gripper state, a one-hot embodiment identifier, and a task embedding.
Task embeddings are derived from CLIP~\cite{radford2021CLIP} given short task descriptions; the language serves only as an identifier, does not inject a linguistic prior, and is similar to task-specific learnable tokens.
Seeker fuses the task embedding with the remaining low-dimensional signals to form its conditioning context (the query-initialization equation in Sec.~\ref{sec:seeker_architecture}), while the policy head concatenates the low-dimensional inputs with a down-projected task embedding.
In multi-embodiment training, ROI localization degrades without the embodiment identifier, so we include it in both Seeker’s conditioning context and the policy head.
For Seeker, we use only the 3D translation and gripper state, since end-effector rotation is largely redundant for coarse ROI localization and can hinder convergence; rotation can be added back for tasks where wrist orientation is essential.
Dataset \textbf{images} are stored at $240 \times 240$ and randomly cropped to $224 \times 224$ during training, which is important for stable policy learning~\cite{mandlekar_2021_robomimic}.
During \textbf{mask extraction}, we drop the two lowest-scored attention heads, renormalize the remaining weights to sum to one, and use nucleus sampling ($\texttt{top\_p}=0.8$) in both simulation and real-robot experiments to suppress low-probability patches.
This hard dropping is applied only at extraction time; during Seeker training, head weights remain soft to preserve differentiability.

\paragraph{Hyperparameters.}
Seeker uses the same hyperparameters as the default Diffusion Policy~\cite{chi2023diffusionpolicy}.
The learning-rate warmup used for Diffusion Policy is also important for stable ROI emergence, as initial updates without warmup can impose strong early attention biases and hinder convergence.

\paragraph{Pixel-level Perturbation.}
\label{app:training:random_overlay}
\begin{figure}[h]
  \centering
  \includegraphics[width=0.55\linewidth]{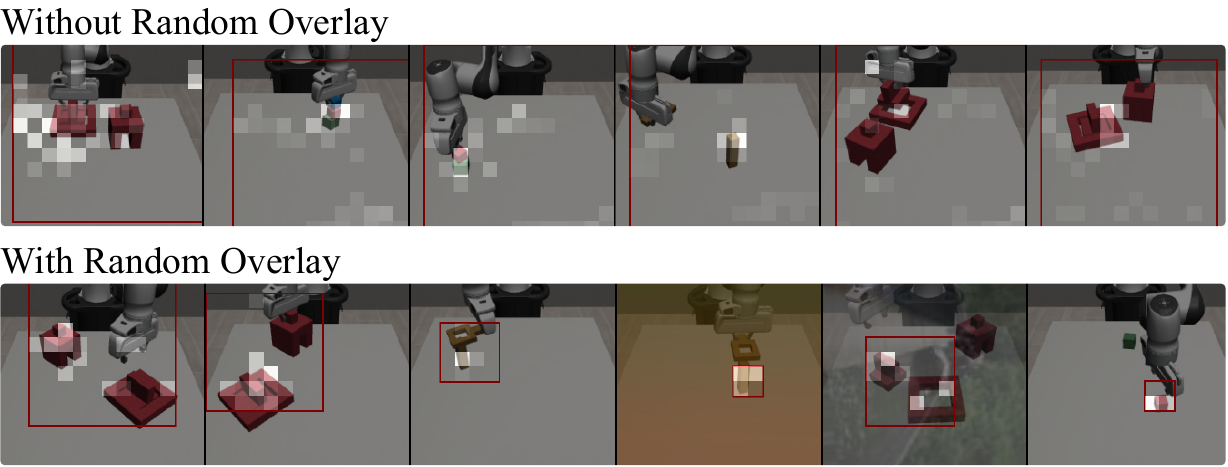}
  \vspace{-1mm}
  \caption{\textbf{Effectiveness of Pixel Perturbation.} At early training stages (10k updates), Seeker trained without Random Overlay exhibits noisy attention on spurious visual artifacts.}
  \label{fig:random_overlay}
  \vspace{-3mm}
\end{figure}

We find pixel-space perturbations crucial for the quality of Seeker's ROI (Fig.~\ref{fig:random_overlay}).
We therefore apply a simple yet effective Random Overlay augmentation, largely following MirrorAug~\cite{mirrorduo2025}.
At each update, each sample is independently overlaid with probability $p_{\mathrm{ov}}$ (i.e., $p_{\mathrm{ov}}$ is the per-sample augmentation probability).
MirrorAug ramps $p_{\mathrm{ov}}$ linearly from $0$ to $0.5$ during a warmup phase.
In contrast, we set the warmup to zero and use $p_{\mathrm{ov}}=0.5$ from the start of training, so that on average half of the samples in every batch are augmented.
Empirically, gradual ramp-up weakens the regularization effect when early spurious biases form, reducing its ability to prevent attention collapse.
The same perturbation is also applied when training the RVT-2-style heatmap predictor.
We mix the original image $I$ with a random texture image $I_{\mathrm{tex}}$ using overlay strength $\alpha=0.6$:
\[
\tilde{I} = \alpha\, I + (1-\alpha)\, I_\mathrm{tex}.
\]

\paragraph{View Complementarity and Sequential Training.}
In multi-view settings (third-person and eye-in-hand, with proprioception), the most informative view can change over the course of an episode.
In particular, the eye-in-hand view often becomes dominant during the interaction phase, while the third-person view can become less informative.
Under joint multi-view training, we observe that the third-person ROI may occasionally shift toward the next target or other salient regions.
This behavior reflects Seeker exploiting cross-view complementarity, rather than a failure of localization.

However, when a standalone third-person ROI is required (e.g., third-person-only policy or cropping for point clouds), this implicit cross-view dependency is no longer available and can lead to degraded performance.
We therefore incorporate additional views sequentially during training.
We first train Seeker using only the third-person view and cache the resulting third-person context features.
We then initialize a multi-view Seeker that takes the cached third-person features while learning to incorporate the additional view(s).
This yields the Seeker variant used in our simulation and real-robot experiments, encourages the third-person representation to remain reliable throughout an episode, and allows adding new cameras without retraining previously incorporated views.

\paragraph{Seeker Localizes Rather Than Controls.}
Although Seeker’s ROI localization emerges from the policy objective, its pooled context is not intended to serve as a standalone action representation.
Directly rolling out actions from Seeker’s context achieves near-zero average success across tasks, with Stack-Three being the highest at 20\%, indicating that Seeker captures mainly \emph{where} to attend while discarding fine-grained geometric and contact cues needed for control.
Fine-tuning DINO improves this context but remains non-competitive with full policy baselines and weakens the pretrained geometric structure used for stable attention, so we keep Seeker frozen and use it only as an explainable visual bottleneck.

\paragraph{Training Objectives.}
\label{app:training:objective_ablation}

\begin{wrapfigure}{r}{0.48\linewidth}
    \vspace{-4.5mm}
    \centering
    \includegraphics[width=\linewidth]{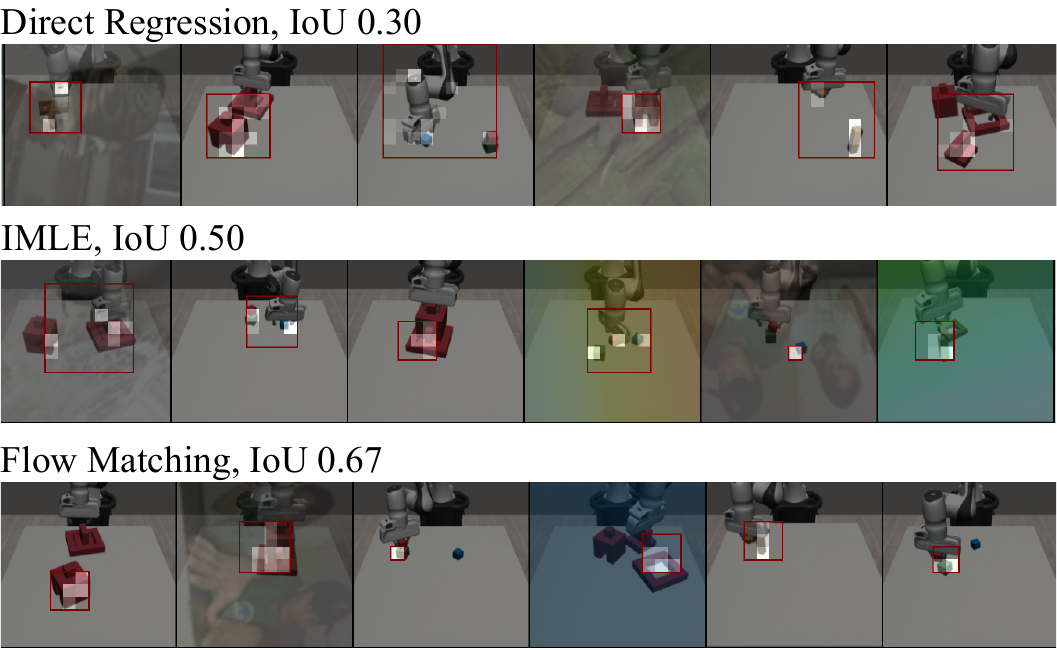}
    \vspace{-4mm}
    \caption{\textbf{Seeker ROI under Different Training Objectives.} Visualizations are shown with the mean IoU to the diffusion-trained Seeker.}
    \label{fig:seeker_training_objectives}
    \vspace{-5mm}
\end{wrapfigure}
We compare Seeker trained with different action objectives, including behavioral cloning (BC, direct regression), flow matching~\cite{lipman_2023_flow}, and Implicit Maximum Likelihood Estimation (IMLE)~\cite{rana2025imle}.
As shown in Fig.~\ref{fig:seeker_training_objectives}, all objectives bias attention toward foreground regions, but produce different ROI structures.
BC yields heavier-tailed ROIs with more scattered noise, likely due to pointwise regression under multimodal expert behavior.
IMLE produces cleaner maps but often spreads attention over multiple plausible foreground cues.
Flow matching yields ROIs closest to the default diffusion-trained Seeker, consistent with its closely related generative formulation.

We quantify these differences by measuring ROI IoU against the diffusion-trained Seeker on 300 trajectories across Square, Stack Three, and Three-Piece Assembly, with 100 trajectories per task.
BC, IMLE, and flow matching achieve mean IoUs of $0.30$, $0.50$, and $0.67$, respectively, matching the qualitative trend in Fig.~\ref{fig:seeker_training_objectives}.

\section{Simulation Task Overviews}
\label{app:task_setup}

We use six simulation tasks from MimicGen~\cite{mandlekar2023mimicgen} and the corresponding released datasets.
We re-render the trajectories at higher image resolution (from $84 \times 84$ to $240 \times 240$) before the $224 \times 224$ random crop used for Seeker.
Five tasks use a Franka Panda arm and one uses a Sawyer arm (Pick and Place D0).
All tasks share a 7D action space, consisting of 3D translation, 3D rotation, and a 1D gripper command.
Each task provides a third-person (agentview) camera and an eye-in-hand camera.
Task descriptions are summarized below.

\noindent\textbf{Full Workspace Shuffling.}
Objects are randomized (both rotation and translation) across the workspace.
\textit{Stack Three D1}: 6 subtasks (3 pick-and-place). Sequentially stack three cubes.
\textit{Square D2}: 2 subtasks (pick-and-insert). Insert a square nut onto a square peg.
\textit{Three-Piece Assembly D2}: 6 subtasks (3 pick-and-place). Sequentially assemble three pieces, requiring precise orientation and placement.
\textit{Threading D2}: 2 subtasks. Pick a needle by its cubic handle and insert the tip into a hole on a fixture.

\noindent\textbf{Constrained Object Shuffling.}
Each component is randomized within a constrained region.
\textit{Coffee Preparation D1}: 7 subtasks. Pick up the mug, place it on the coffee machine, open the capsule slot's lid, open the drawer, pick up a capsule, place it into the slot, and close the lid. The drawer has no spatial variation.
\textit{Pick and Place D0}: 8 subtasks. Sequentially place four objects into their corresponding baskets. Objects start on the left, and the target baskets are fixed on the right.
\section{Evaluation with 1,000 Demonstrations}
\label{app:experiments:high_data}

We increase the downstream policy training budget to 1,000 demonstrations per task while reusing the same frozen Seeker trained with 100 demonstrations per task.

\subsection{RGB Policy Learning.}
Tab.~\ref{tab:sim_high_data} shows that Seeker retains an advantage over MirrorAug on five of six tasks, achieving $77.9\%$ average success compared with $70.2\%$.
This suggests that additional data does not substitute for the benefits brought by an explicit spatial bottleneck.

\begin{table}[ht]
    \centering
    \footnotesize
    \setlength{\tabcolsep}{3pt}
    \begin{tabular}{lccccccc}
    \toprule
    Method & \textbf{Stack Three} & \textbf{Square} & \textbf{3-P Assembly}
    & \textbf{Coffee Prep.} & \textbf{Pick \& Place} & \textbf{Threading} & Avg. \\
    \midrule
    MirrorAug
    & \textbf{96.0} & 59.3 & 65.3 & 82.7 & 62.7 & 55.3 & 70.2 \\
    Seeker
    & 94.7 & \textbf{69.3} & \textbf{74.0}
    & \textbf{84.7} & \textbf{84.0} & \textbf{60.7} & \textbf{77.9} \\
    \bottomrule
    \end{tabular}
    \vspace{1mm}
    \caption{
    \textbf{RGB Policy with 1,000 Demonstrations per Task.}
    Success rate (\%) averaged over three seeds.
    }
    \label{tab:sim_high_data}
    \vspace{-4mm}
\end{table}

\subsection{Background Generalization.}
We also evaluate the larger-data policies under shuffled tabletop textures, using either augmentation alone or augmentation combined with background-randomized training.
Tab.~\ref{tab:sim_high_data_ood} reports absolute OOD success and relative decreases from the original-background ID results.
With augmentation alone, Seeker achieves higher OOD success and smaller relative decreases on all three tasks.
At this data budget, background randomization improves OOD success for both methods across all three tasks, unlike its mixed effect in the 100-demonstration setting.
These results are consistent with the explanation that learning to separate task-relevant structure from complex textures benefits from additional data.
Seeker remains stronger under randomized training; with augmentation alone, it also exceeds randomized MirrorAug on Square and Three-Piece Assembly, while performing slightly below it on Stack Three.

\begin{table}[ht]
    \centering
    \footnotesize
    \setlength{\tabcolsep}{5pt}
    \begin{tabular}{llccc}
    \toprule
    Training & Method & \textbf{Stack Three} & \textbf{Square} & \textbf{3-P Assembly} \\
    \midrule
    \multirow{2}{*}{Augmentation only}
    & MirrorAug
    & 56.7 [40.9\%] & 26.7 [55.0\%] & 30.7 [53.0\%] \\
    & Seeker
    & \textbf{79.3} [16.3\%]
    & \textbf{56.0} [19.2\%]
    & \textbf{53.3} [28.0\%] \\
    \midrule
    \multirow{2}{*}{Domain randomization}
    & MirrorAug
    & 80.7 [15.9\%] & 37.3 [37.1\%] & 45.3 [30.6\%] \\
    & Seeker
    & \textbf{88.0} [7.1\%]
    & \textbf{60.0} [13.4\%]
    & \textbf{68.0} [8.1\%] \\
    \bottomrule
    \end{tabular}
    \vspace{1mm}
    \caption{
    \textbf{Background Generalization with 1,000 Demonstrations per Task.}
    OOD success (\%) averaged over three seeds.
    Brackets show relative decreases from the corresponding
    original-background ID scores in Tab.~\ref{tab:sim_high_data},
    calculated as $100(s_{\mathrm{ID}}-s_{\mathrm{OOD}})/s_{\mathrm{ID}}$.
    }
    \label{tab:sim_high_data_ood}
    \vspace{-4mm}
\end{table}
\section{Data Efficiency and Spatial Coverage}
\label{app:spatial_coverage}
\begin{figure}[h]
    \vspace{-5mm}
    \centering
    \subfloat[Stack Three: green cube\label{fig:spatial_coverage_stack_three}]{
        \includegraphics[width=0.31\linewidth]{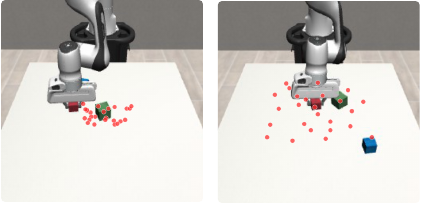}
    }
    \hfill
    \subfloat[Square: peg\label{fig:spatial_coverage_square}]{
        \includegraphics[width=0.31\linewidth]{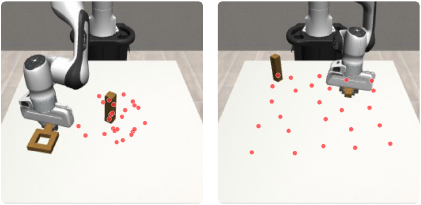}
    }
    \hfill
    \subfloat[3-Piece Assembly: first piece\label{fig:spatial_coverage_three_piece}]{
        \includegraphics[width=0.31\linewidth]{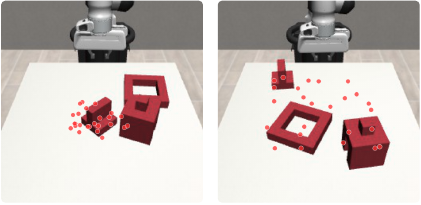}
    }
    \vspace{-1mm}
    \caption{
    \textbf{Sub-task Spatial Coverage.}
    Sub-task target distributions used to construct coverage-controlled 25-demo equivalent Seeker pretraining subsets.
    Each panel shows one representative sub-task, comparing the least diverse subset on the left with the most diverse subset on the right.
    }
    \label{fig:app:spatial_coverage}
    \vspace{-1mm}
\end{figure}

This section contextualizes Seeker's training overhead with a task-specific coverage diagnostic.
Our main experiments train Seeker on the full 100-demo policy-training set, giving a conservative estimate of total training cost.
Here, we ask whether this cost can be reduced when the selected demonstrations still preserve the spatial variation needed for ROI learning.

We train task-specific Seekers for the coverage-selected subsets and compare them with Full-100, the unchanged multi-task Seeker from the main experiments, trained on all 100 demonstrations per task without coverage selection.
In a multi-task Seeker, the effective data requirement is shaped not only by per-task coverage, but also by task balance and inter-task interference, especially when the pretraining set is reduced.
Our early experiments showed that multi-task Seeker needed more data to match isolated task-specific Seekers.
Comparisons among the task-specific variants probe spatial coverage; comparisons with Full-100 also change the training regime and do not isolate coverage alone.

We construct coverage-controlled subsets using privileged sub-task annotations only for selection.
Each task is decomposed into sub-tasks, and demonstrations within each sub-task are ranked by farthest-point sampling over target poses.
An $N$-demo equivalent setting denotes the top or bottom $N$ demonstrations selected independently per sub-task from the 100-demo pool, rather than $N$ trajectories selected globally.
Seeker is trained on the corresponding demonstration windows, with a fixed-range normalizer to avoid subset-specific normalization shifts.

Fig.~\ref{fig:app:spatial_coverage} visualizes the least and most diverse 25-demo equivalent subsets for representative sub-tasks.
Tab.~\ref{tab:app:spatial_coverage} shows that task-specific Seeker is robust to reduced pretraining data when spatial coverage is preserved: Most-25 is within $0.4$ points of Full-100 on average, and Most-50 matches Full-100.
Least-25 remains competitive but drops by $3.6$ points, mainly on Square and 3-Piece Assembly.
This suggests that spatial distribution matters most when the pretraining budget is small, especially for tasks such as 3-Piece Assembly where similarly colored, shape-dependent objects induce broader pose-dependent contact regions.

\providecommand{\pmstd}[2]{#1{\scriptsize$\pm#2$}}

\begin{table}[h]
\centering
\footnotesize
\setlength{\tabcolsep}{4pt}
\begin{tabular}{lcccc}
\toprule
\textbf{Coverage}
& \textbf{Stack Three D1}
& \textbf{Square D2}
& \textbf{3-Piece Assembly D2}
& \textbf{Avg.} \\
\midrule
\textbf{Full-100}
& \textbf{\pmstd{81.3}{1.2}}
& \textbf{\pmstd{46.0}{7.2}}
& \textbf{\pmstd{58.7}{2.3}}
& \textbf{62.0} \\
\cmidrule(lr){2-5}
Least-25
& \pmstd{81.3}{1.2}
& \pmstd{40.7}{1.2}
& \pmstd{53.3}{2.3}
& $58.4_{\gaindown{3.6}}$ \\
Most-25
& \pmstd{78.7}{4.2}
& \pmstd{44.7}{3.1}
& \pmstd{61.3}{2.3}
& $61.6_{\gaindown{0.4}}$ \\
Most-50
& \pmstd{80.0}{2.0}
& \pmstd{45.3}{7.6}
& \pmstd{60.7}{5.0}
& $62.0$ \\
\bottomrule
\end{tabular}
\vspace{1mm}
\caption{
\textbf{Effect of Seeker Pretraining Spatial Coverage.}
Success rates (\%) are computed as the mean over three seed-level best checkpoints.
Full-100 is the default multi-task Seeker from the main experiments, with all 100 demonstrations per task and no coverage selection.
The task-specific variants Least-25 and Most-25 use the least and most spatially diverse 25-demo equivalent subsets, respectively; Most-50 uses the most spatially diverse 50-demo equivalent subset.
Average subscripts report the change relative to Full-100.
}
\label{tab:app:spatial_coverage}
\vspace{-3mm}
\end{table}

Overall, this diagnostic suggests that task-specific Seeker pretraining can be substantially reduced with spatially covered subsets.
In the multi-task setting, however, the reducible overhead depends on preserving both spatial coverage and task balance while mitigating cross-task interference. Quantifying this trade-off is beyond the scope of this diagnostic.

\section{Experimental Details and Results}
\label{app:experimental_details}
\subsection{RGB Cropping}
\label{app:experiments:full_results}
Tab.~\ref{app:tab:main_sim_std} reports the full RGB results with standard deviations.

\begin{table}[ht]
    \centering
\footnotesize

\providecommand{\pmstd}[2]{#1{\scriptsize$\pm#2$}}

\setlength{\tabcolsep}{2.6pt}

\begin{tabular}{l *{6}{c}}
\toprule
&
\textbf{Stack Three} &
\textbf{Square} &
\textbf{3-P Assembly} &
\textbf{Coffee Prep.} &
\textbf{Pick \& Place}$^\dagger$ &
\textbf{Threading} \\
\midrule

& \multicolumn{6}{c}{\textit{Input-level Comparison}} \\
\cmidrule(lr){2-7}

DiffPo (Pre) &
\pmstd{52.7}{5.4} &
\pmstd{22.0}{3.5} &
\pmstd{14.0}{9.2} &
\pmstd{65.3}{2.1} &
\pmstd{20.0}{5.3} &
\pmstd{20.7}{2.1} \\

MirrorAug &
\pmstd{68.0}{3.3} &
\pmstd{32.7}{2.5} &
\pmstd{20.7}{1.2} &
\pmstd{62.0}{2.0} &
\pmstd{20.7}{3.1} &
\pmstd{18.7}{1.2} \\

RVT2-Crop &
\pmstd{71.3}{1.2} &
\pmstd{26.0}{2.0} & 
\pmstd{26.7}{1.2} & 
\pmstd{69.3}{2.3} & 
\pmstd{49.3}{3.1} & 
\pmstd{12.7}{1.2} \\

\addlinespace[0.5pt]

Seeker (ours) &
\pmstd{\textbf{81.3}}{1.2} &
\pmstd{\textbf{46.0}}{7.2} &
\pmstd{\textbf{58.7}}{2.3} &
\pmstd{\textbf{82.0}}{2.0} &
\pmstd{\textbf{69.3}}{1.2} &
\pmstd{\textbf{38.0}}{2.0} \\
\midrule

& \multicolumn{6}{c}{\textit{External SOTA and Privileged ROI References}} \\
\cmidrule(lr){2-7}

RAVEN &
\pmstd{80.7}{2.1} &
\pmstd{50.0}{6.9} &
\pmstd{26.7}{7.3} &
\pmstd{72.7}{7.3} &
\pmstd{54.7}{1.2} &
\pmstd{28.0}{3.5} \\

Oracle ROI$^\ast$ &
\pmstd{80.7}{1.2} &
\pmstd{44.7}{4.2} &
\pmstd{62.7}{1.2} &
\pmstd{81.3}{4.2} &
\pmstd{71.3}{1.2} &
\pmstd{44.7}{1.2} \\

\bottomrule
\end{tabular}

    \vspace{1mm}
    \caption{
    \textbf{Data Efficiency on MimicGen}~\cite{mandlekar2023mimicgen}.
    Success rate (\%) on six tasks with 100 demonstrations, averaged over three seeds and reported as mean{\scriptsize$\pm$std}.
    RAVEN and DiffPo (Pre) results are taken from the RAVEN paper~\cite{klee2026raven} where protocol-matched.
    $^\dagger$Pick \& Place uses our strict episode-success reruns. $^\ast$The oracle provides stage-aware affordance ROIs.
    }
    \label{app:tab:main_sim_std}
    \vspace{-4mm}
\end{table}

\subsection{Point-cloud Filtering}
\label{app:experiments:pointcloud}


\paragraph{Implementation.}
For each observation, we unproject calibrated depth images from all cameras into a shared world-frame point cloud.
Given a Seeker bounding box from camera $\tilde{c}$, we reproject all world-frame points into the image plane of $\tilde{c}$ and retain only points whose projected pixel coordinates fall inside the box.
This yields the Seeker-cropped point cloud while still leveraging points from all cameras.
Unlike cropping each camera image independently before 3D projection, this strategy uses a single Seeker prediction and does not require training camera-specific Seekers.
For the remaining point-cloud experiments, we follow EquiDiff's~\cite{wang2024equivariant} default DP3~\cite{ze20243ddif} setup.

\begin{table}[ht]
    \centering
    \footnotesize
\centering
\begin{tabular}{lcccc}
    \toprule
    \textit{Manual}      & \xmark & \cmark & \xmark & \cmark \\
    \textit{Seeker}      & \xmark & \xmark & \cmark & \cmark \\
    \midrule
    \textbf{Stack Three D1}
        & $0.0_{\pm 0.0}$ & $23.3_{\pm 1.1}$ & $34.6_{\pm 5.7}$ & $\textbf{47.3}_{\pm 8.0}$ \\
    \textbf{Square D2} 
        & $0.0_{\pm 0.0}$ & $6.0_{\pm 1.1}$ & $\textbf{8.6}_{\pm 1.1}$ & $7.3_{\pm 1.1}$ \\
    \textbf{3-Piece Assem.\ D2}
        & $0.0_{\pm 0.0}$ & $1.3_{\pm 1.1}$ & $\textbf{6.0}_{\pm 2.0}$ & $\textbf{6.0}_{\pm 0.0}$ \\
    \textbf{Coffee Prep.\ D1}  
        & $0.0_{\pm 0.0}$ & $22.0_{\pm 6.9}$ & $42.0_{\pm 6.0}$ & $\textbf{64.3}_{\pm 8.0}$ \\
    \textbf{Pick \& Place D0}  
        & $0.0_{\pm 0.0}$ & $1.3_{\pm 1.1}$ & $12.6_{\pm 1.1}$ & $\textbf{26.0}_{\pm 3.5}$ \\
    \textbf{Threading D2}
        & $0.0_{\pm0.0}$ & $18.0_{\pm 2.0}$ & $22.6_{\pm 4.6}$ & $\textbf{28.6}_{\pm 4.2}$ \\
    \bottomrule
\end{tabular}
    \vspace{1mm}
    \caption{\textbf{Full Point-cloud Results}. We compare different point-cloud cropping strategies across all tasks over three seeds, reporting the mean and standard deviation.}
    \label{tab:app:pointcloud_std}
    \vspace{-4mm}
\end{table}

\begin{figure}[ht]
    \centering
    \subfloat[Rotation\label{fig:app:pointcloud_failures:1}]{
        \includegraphics[width=0.3\linewidth]{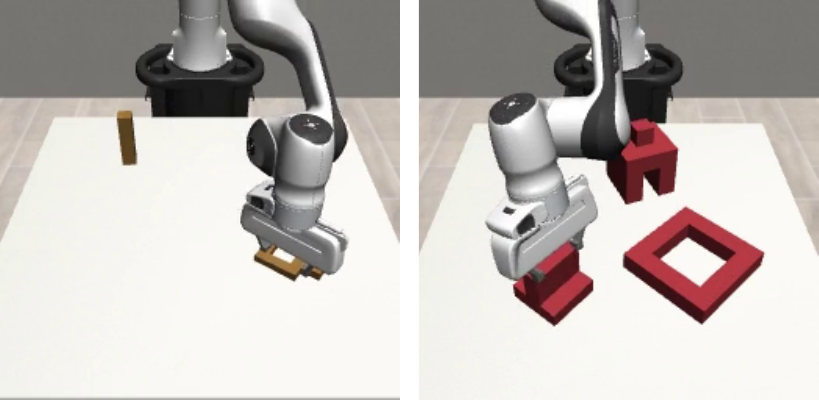}
    }
    \hspace{4mm}
    \subfloat[Workspace\label{fig:app:pointcloud_failures:2}]{
        \includegraphics[width=0.3\linewidth]{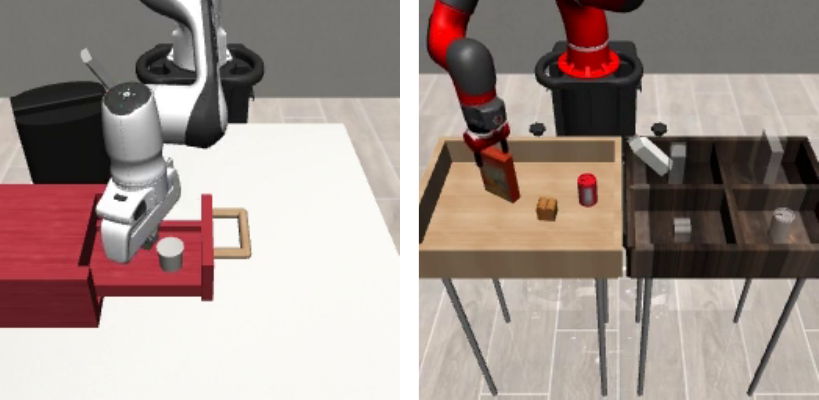}
    }
    \vspace{-1mm}
    \caption{\textbf{Point-cloud Policy Stress Cases.}
    Under combined cropping, failures mainly arise from (a) inaccurate orientation estimation and (b) sparse task-relevant points in large workspaces, making fine-grained motions difficult to infer.}
    \label{fig:app:pointcloud_failures}
    \vspace{-4mm}
\end{figure}

\paragraph{Failure Analysis.}
Tab.~\ref{tab:app:pointcloud_std} reports full results across all tasks and seeds.
Seeker plus manual workspace cropping achieves the highest success on four tasks and ties for highest on a fifth, but point-cloud policies still lag image-based policies on most tasks.
Fig.~\ref{fig:app:pointcloud_failures} shows two typical failure modes.
First, DP3 can struggle with fine-grained rotation inference, so Seeker's spatial focus does not fully resolve orientation ambiguity in the point-cloud representation.
This is most pronounced in Three-Piece Assembly, which has broad up-axis rotation variation, and less severe in Coffee Preparation, where orientation variation is smaller.
Second, in large workspaces such as Pick \& Place, task-relevant points can remain sparse after cropping, especially when Seeker under-crops and leaves distractor points.

\subsection{Random Overlay and Guided Overlay}
\label{app:experiments:guided_aug}
In Sec.~\ref{sec:simulation}, we introduced Random Overlay and Seeker's guided augmentation. We detail them below:

\textbf{Random Overlay} is applied by blending the input image $I$ with a randomly sampled background $I_{\text{tex}}$,
{\small \(
\tilde I = \alpha I + (1-\alpha) I_{\text{tex}},
\)}
where $\alpha \in [0,1]$ is the constant original-image preservation coefficient.
For downstream policies, the augmentation is applied with probability $p$, linearly warmed up to 0.5 over the first 2500 training steps, with $\alpha$ fixed to 0.6.

\textbf{Guided Overlay.}
Guided Overlay follows the same procedure, but replaces the global blending coefficient with a spatially varying map derived from Seeker’s mask $M$.
The mask is produced at patch resolution (e.g., $14 \times 14$ for a $224 \times 224$ input), and we bilinearly upsample it to the image resolution before blending.
The image $I$ is either the ROI-cropped third-person view or the full eye-in-hand view.
We convert $M$ (summing to one) into an image-valued map by $M_{\text{norm}} = M / \max(M)$ and clamp it to $[0.3, 0.8]$ to obtain $M_{\text{clamp}}$.
We then blend per pixel as
\[
\tilde I = M_{\text{clamp}} \odot I + (1 - M_{\text{clamp}}) \odot I_{\text{tex}}.
\]
Clamping accounts for the mask’s coarse patch resolution and imperfect localization.
Overly extreme blending that fully replaces low-confidence regions or preserves high-confidence regions can degrade performance.
Clamping moderates these extremes: low-confidence regions are strongly perturbed but still partially preserved, while high-confidence regions are only mildly perturbed rather than fully protected.

\section{Real Experiments}
\label{app:real_world_exp}

\subsection{Tasks and Setup}
\label{app:real_world_exp-task_description}

All tasks are performed on a UFactory xArm~7 with fixed third-person and eye-in-hand RGB cameras.
Demonstrations are collected via teleoperation, and objects are randomly shuffled within the workspace during both data collection and policy evaluation.
All tasks use position-based end-effector control at 10~Hz.
Both proprioceptive states and actions are 10-dimensional, comprising a 3D Cartesian end-effector position, a 6D representation of end-effector rotation, and a 1D continuous gripper state.
Image observations are captured at $848 \times 480$, center-cropped for the third-person view and square-padded for the eye-in-hand view, then resized to $240 \times 240$.
Task descriptions are provided below, with dataset statistics summarized in Tab.~\ref{app:tab:real_task_summary}.

\textbf{Coffee Transport} requires the robot to scoop coffee beans from a yellow bowl using a spoon and pour them into an espresso cup.
This task requires precise control of tool orientation and pouring motion, with the small white espresso cup placed against a white-dominant textured background, making target localization brittle from RGB observations.
Successful execution requires coordinated reaching, scooping, transporting, and controlled tilting without spilling.

\textbf{Table Cleanup} requires the robot to sequentially grasp and insert two toy vegetables with different geometries, a long thin carrot and a round tomato, into a bin, followed by placing a lid on top.
The task involves multiple object interactions, ordered subtasks, and non-trivial 3D rotations during insertion, requiring temporal consistency and accurate pose control across subtasks.

\textbf{Board Assembly} requires the robot to sequentially pick three pieces with distinct base shapes, round, square, then hexagonal, and insert them into corresponding slots on a board.
Although the geometries match exactly, the part colors intentionally do not correspond to the slot colors, increasing the reasoning difficulty.
Successful execution requires millimeter-level accuracy and robust shape-based pose estimation across a long horizon.

\begin{table}[ht]
    \centering
    \footnotesize
    \setlength{\tabcolsep}{4pt}
    \begin{tabular}{@{}lccccc@{}}
        \toprule
        Task & \# Demos & \# Subtasks & ImgD & PropD & ActD \\
        \midrule
        \textbf{Coffee Transport} & 50  & 4 & $2 \times 3 \times 240 \times 240$ & 10 & 10 \\
        \textbf{Table Cleanup}    & 100 & 6 & $2 \times 3 \times 240 \times 240$ & 10 & 10 \\
        \textbf{Board Assembly}   & 100 & 6 & $2 \times 3 \times 240 \times 240$ & 10 & 10 \\
        \bottomrule
    \end{tabular}
    \vspace{1mm}
    \caption{\textbf{Real-world task summary.}
    \# Demos: demonstrations used for both Seeker and downstream policies; ImgD: preprocessed RGB observation dimensionality; PropD and ActD: proprioception and action dimensions.}
    \label{app:tab:real_task_summary}
    \vspace{-4mm}
\end{table}

\subsection{Training Details}
\label{app:real_world_exp-training}

We train our methods and all baselines on real-world data using the same protocol as in simulation.
Limited demonstrations and complex visual backgrounds make convergence more challenging, especially with image-space noise from Random Overlay for baselines and Guided Overlay for Seeker policies.
We therefore adjust two hyperparameters: the overlay warm-up is increased to 5{,}000 steps for all policies, and Seeker training is extended to 600 epochs.

\subsection{Seeker Visualization}
\label{app:real_world_exp-visualization}

\begin{figure}[!tbp]
    \centering

    \subfloat[\textit{Coffee Transport}\label{fig:real_keyframe_coffee}]{
        \includegraphics[width=0.6\linewidth]{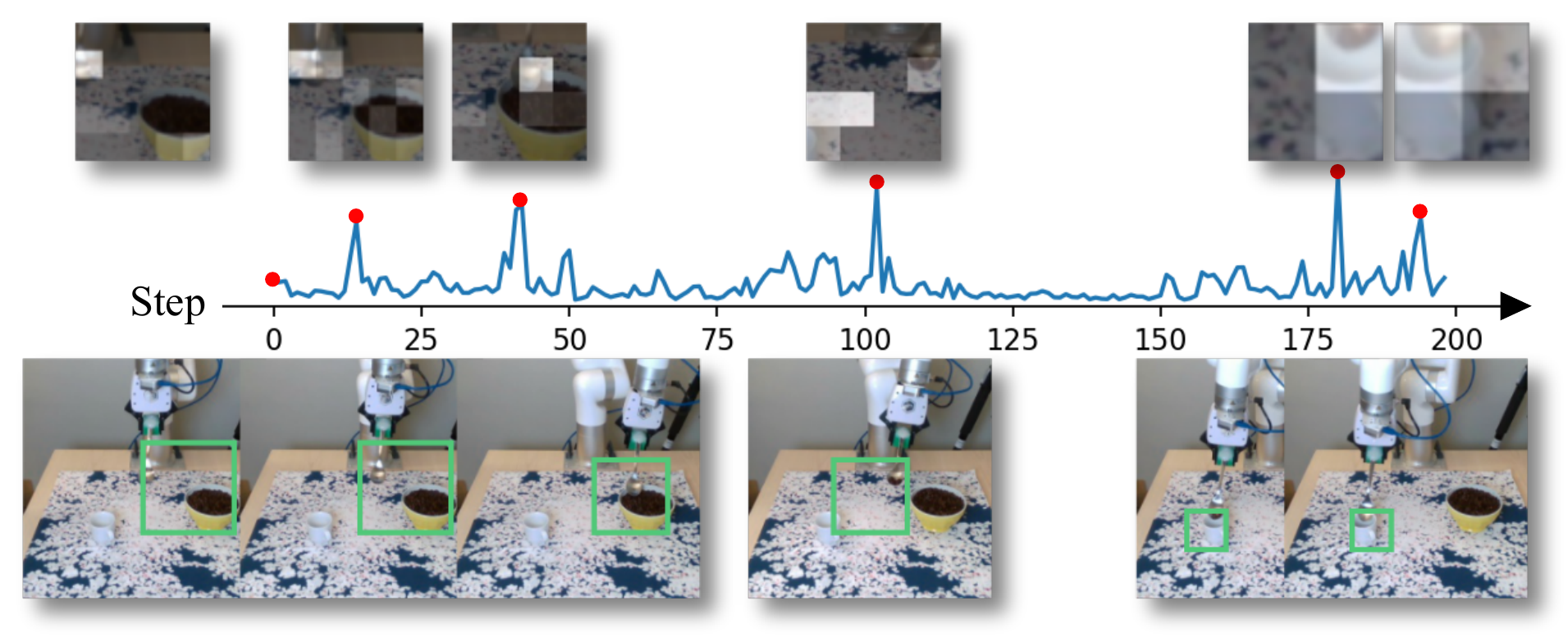}
    }
    \vspace{-2mm}
    \\

    \subfloat[\textit{Table Cleanup}\label{fig:real_keyframe_cleanup}]{
        \includegraphics[width=0.48\linewidth]{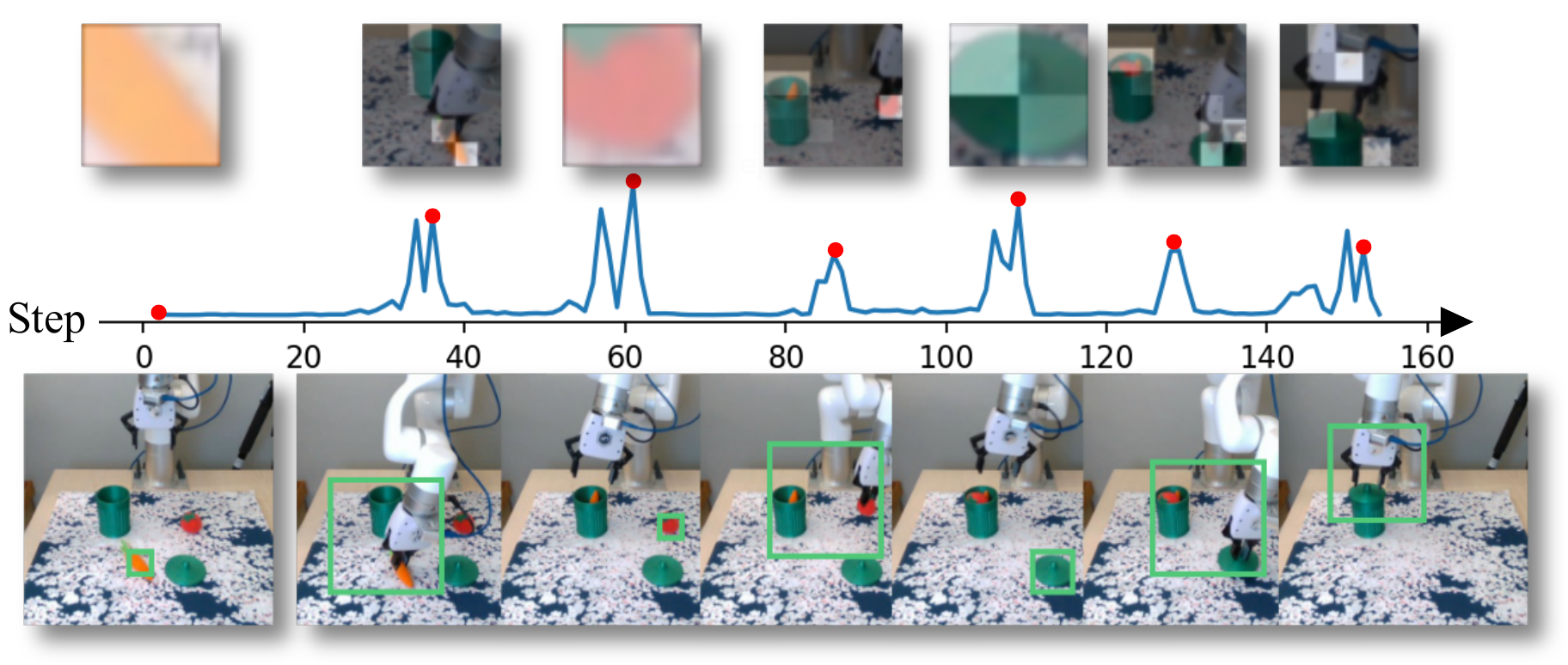}
    }
    \hfill
    \subfloat[\textit{Board Assembly}\label{fig:real_keyframe_assembly}]{
        \includegraphics[width=0.48\linewidth]{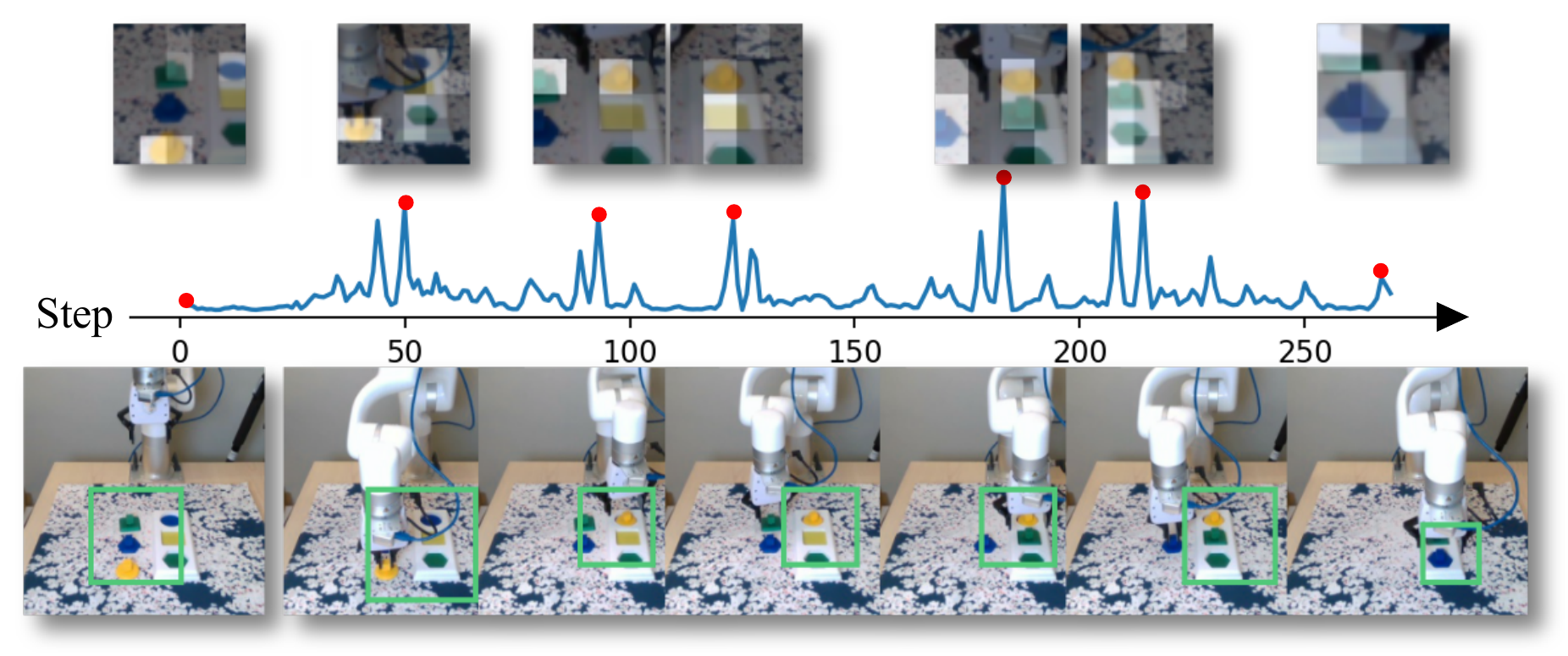}
    }

    \vspace{-2mm}
    \caption{\textbf{Seeker-selected Keyframes for Real-world Tasks.}
    Predicted ROIs, crops, and masks are aligned with peaks in pairwise cosine distance from Seeker's context feature.}
    \label{app:fig:real_keyframes}
    \vspace{-4mm}
\end{figure}

Fig.~\ref{app:fig:real_keyframes} shows additional real-world visualizations, including Seeker-selected keyframes, masks, and crop outputs.
Keyframe scores are computed from the cosine distance between consecutive Seeker context vectors, as in simulation, and highlight meaningful stage transitions.
Qualitatively, Seeker maintains task-relevant localization and progression-aware focus switching on real hardware.

Compared with MimicGen~\cite{mandlekar2023mimicgen}, real-world demonstrations are generated by human operators and contain smoother continuous translation and rotation within each subtask.
Accordingly, Seeker more often exhibits an \textit{interaction-centric} focus, covering both the manipulated object and the target, such as the spoon tip and cup in \textit{Coffee Transport} or the part and slot in \textit{Board Assembly}.
In \textit{Table Cleanup}, where manipulation involves less rotation, Seeker more often follows an \textit{object-centric} pattern.
Overall, these results suggest that Seeker can switch between \textit{object-centric} and \textit{interaction-centric} bottlenecks depending on the task and data distribution.





\end{document}